\pdfoutput=1
\documentclass[10pt,letterpaper,onecolumn]{article}

\usepackage{preprint,times}
\usepackage{float}

\usepackage{amsmath,amssymb,mathtools}
\usepackage{graphicx}
\usepackage{booktabs}
\usepackage{tabularx}
\usepackage{longtable}
\usepackage{multirow}
\usepackage{url}
\usepackage[table]{xcolor}
\definecolor{besttarget}{RGB}{230,238,244}
\usepackage{algorithm}
\usepackage[noend]{algorithmic}
\usepackage{placeins}
\usepackage{needspace}
\usepackage[hidelinks]{hyperref}
\hypersetup{pdftitle={HoloCosm: Beyond Single-Metric Benchmark Compression Across Multiple Metrics},
pdfauthor={Yupei Lin, Zhongzhan Huang, Junxin Li, Guoming Ling, Shanshan Zhong, Hefeng Wu, Yukai Shi, Liang Lin}}
\date{}

\newcommand{\appmain}[1]{%
  \textbf{\ref{#1}} &
  \textbf{\nameref{#1}} &
  \textbf{\pageref{#1}} \\[0.25em]
}

\newcommand{\appsub}[1]{%
  \hspace{1.2em}\ref{#1} &
  \hspace{0.3em}\nameref{#1} &
  \pageref{#1} \\
}

\newcommand{\bestcell}[1]{%
  \cellcolor{besttarget}#1%
}

\title{
HoloCosm:
Beyond Single-Metric Benchmark Compression Across Multiple Metrics
}

\author{
Yupei Lin$^{1}$\thanks{These authors contributed equally.},
Zhongzhan Huang$^{1}$\footnotemark[1],
Junxin Li$^{1}$, Guoming Ling$^{1}$,
Shanshan Zhong$^{1}$\\
Hefeng Wu$^{1}$,
Yukai Shi$^{2}$, Liang Lin$^{1}$\thanks{Corresponding author.}\\
\normalfont $^{1}$Sun Yat-sen University\quad
$^{2}$Guangdong University of Technology
}

\begin{document}
\maketitle
\enlargethispage{6pt}

\begin{abstract}
Benchmark compression reduces evaluation cost by selecting a small set
of samples. Existing methods typically guide selection using a single
metric, although benchmarks often assess models along multiple dimensions.
We find that model rankings are often weakly correlated across metrics,
indicating that different metrics capture distinct aspects of model
performance. Samples selected for one metric do not reliably preserve
scores and rankings on others when used for compressed evaluation. At 80\% compression, each of the five
single-metric baselines yields higher reconstruction error than equal-size
random sampling in over 70\% of cross-metric comparisons.
To address this challenge, we present HoloCosm, which jointly minimizes
normalized score-reconstruction error across source models and metrics
while constraining ranking distortion. 
We evaluate HoloCosm across multiple compression rates on
15 language, multimodal, VLA, and agent benchmark configurations
with 85 component metrics. With 80\% compression,
HoloCosm outperforms Random, the strongest baseline overall: average NMAE
drops by 35.3\% to 0.236, worst-component NMAE drops by 39.5\% to 0.330,
and average Spearman correlation rises from 0.884 to 0.915.
In addition, HoloCosm remains competitive with existing baselines in
single-metric evaluation. We will release our code and experimental results.

\end{abstract}

\begingroup
\setlength{\intextsep}{0pt}
\begin{figure}[H]
  \centering
  \makebox[\linewidth][c]{\includegraphics[width=6.5in]{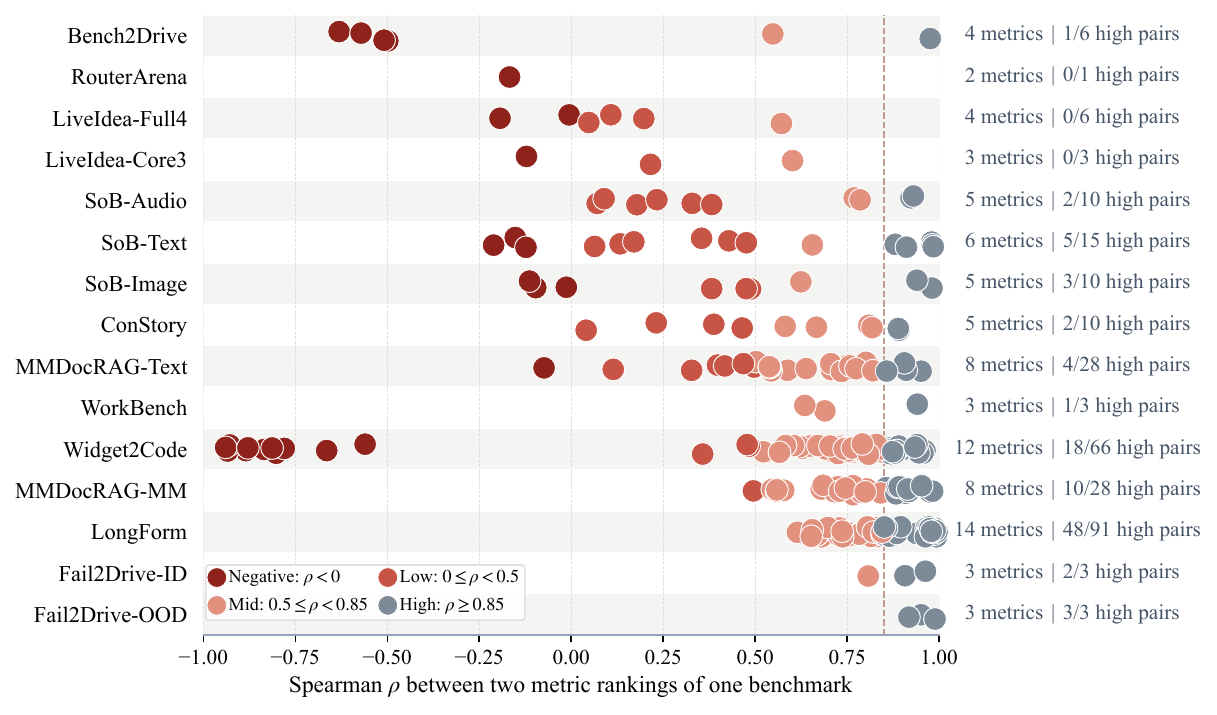}}
  \caption{
  \textbf{Model-ranking consistency across metrics.}
Each point shows Spearman $\rho$ between rankings induced by two official
metrics, all direction-aligned so higher is better. Colors show correlation
ranges, the dashed line marks $\rho\geq0.85$, and right-side annotations give
metric counts and high-correlation pairs. Weak and negative correlations are common.
  }
  \label{fig:motivation}
\end{figure}
\endgroup

\section{Introduction}
\label{sec:introduction}

As the number of models and the scale of evaluation continue to grow, repeatedly evaluating models on full benchmarks incurs increasing computational and time costs. Benchmark compression reduces this burden by selecting a small set of representative samples to approximate full-benchmark model scores and rankings. Existing benchmark compression methods (\textbf{BCMs}), such as AnchorPoints, tinyBenchmarks, mRMR-based compression, SubLIME, and ZipBench, aim to substantially reduce benchmark size while maintaining high evaluation fidelity through representative sample selection for downstream model comparison \citep{vivek2024anchor,polo2024tinybenchmarks,bowyer2026efficient,saranathan2025sublime,huang2026zipbenchlowcostframeworkcompressing}. However, these methods typically guide sample selection using a single evaluation metric, whereas modern benchmarks often assess models along multiple dimensions. This exposes a key issue: samples considered representative under one metric may not be equally representative for other metrics.

We therefore examine the consistency of model rankings across evaluation metrics. Across 85 official metrics from 15 benchmark configurations spanning language, multimodal, VLA, and agent tasks, we compute Spearman correlations between metric-induced model rankings within each benchmark. As shown in Fig.~\ref{fig:motivation}, many metric pairs exhibit weak ranking agreement, with negative correlations appearing across several benchmarks. Using $\rho \geq 0.85$ as the threshold for high correlation \citep{granato2014statistics}, we find that highly correlated pairs are only a small fraction in many benchmarks: Bench2Drive has only one out of six such pairs, while LiveIdea-Full4 \citep{ruan2026evaluating} has none. These results show that different metrics capture distinct, sometimes conflicting, model-comparison information, raising a concern for compression guided by a single metric.

\begin{figure*}[ht]
  \centering
  \includegraphics[width=\textwidth]{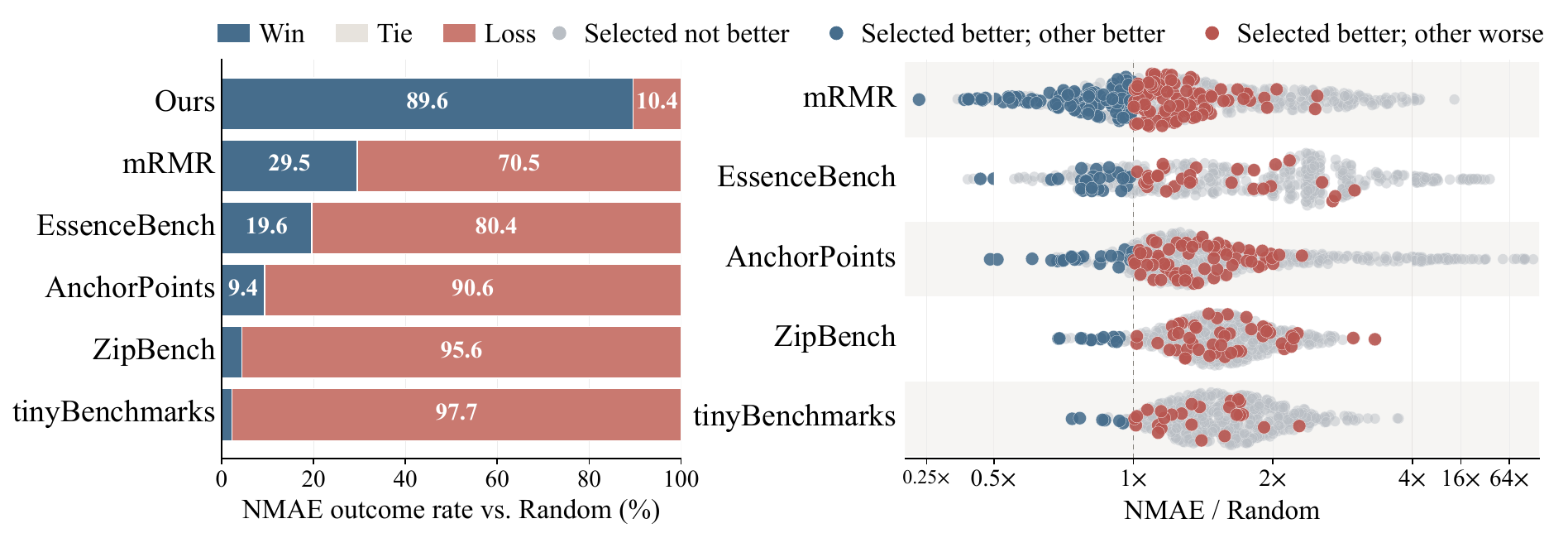}
\caption{
\textbf{Single-metric selection fails across metrics.}
\textbf{Left:} Cross-metric NMAE win, tie, and loss rates against Random across
15 benchmarks at 80\% compression. HoloCosm wins in 89.6\% of comparisons.
\textbf{Right:} NMAE relative to Random on metrics other than
the selection metric, with compressed axis tails.
Red points show cross-metric failure despite selected-metric gains.
}
  \label{fig:challenge}
\end{figure*}

These cross-metric differences raise a question for single-metric compression: can samples selected to represent one metric also preserve the evaluation information captured by others? To examine this question, we select items using one metric and evaluate them on the remaining metrics of the same benchmark, comparing against equal-size Random sampling at 80\% compression. As shown in the left panel of Fig.~\ref{fig:challenge}, existing single-metric BCMs perform worse than Random in 70.5\%--97.7\% of cross-metric score-reconstruction comparisons across five baselines. The right panel further shows that even when the selected samples achieve lower NMAE than Random on the selection metric, this advantage often fails to carry over to other metrics, where reconstruction errors instead exceed those of Random. These results show that a subset representative of one metric does not necessarily preserve the information needed to evaluate others.

To address this systematic cross-metric failure, we present HoloCosm, which directly optimizes evaluation fidelity across multiple metrics rather than assuming that items selected for one metric will transfer to others. 
Using full-benchmark evaluation results from source models, HoloCosm minimizes normalized score-reconstruction error across models and metrics while constraining ranking distortion.
Size-matched reference selections calibrate the ranking-distortion constraint, a feasibility-first exchange search refines the selected items, and a source-assisted correction estimates full-benchmark scores while preserving within-metric rankings. At 80\% compression (20\% retention), HoloCosm achieves the best aggregate performance across 15 benchmark configurations and 85 evaluation metrics. Despite being designed for multi-metric evaluation, HoloCosm remains competitive with existing single-metric BCMs in single-metric settings. Our contributions are:
\begin{itemize}
\setlength{\topsep}{2pt}
\setlength{\itemsep}{2pt}
\item \textbf{Multi-metric inconsistency.}
Across 85 official metrics from 15 benchmark configurations, we find widespread weak and negative correlations between metric-induced model rankings, showing that different evaluation dimensions often capture substantially different information across different types of benchmarks.

\item \textbf{Systematic cross-metric failure.}
At 80\% compression, existing single-metric BCMs perform worse than Random in 70.5\%--97.7\% of cross-metric score-reconstruction comparisons, revealing a systematic failure of single-metric-guided compression. To our knowledge, we are the first to systematically study this multi-metric compression problem.

\item \textbf{Multi-metric benchmark compression.}
We introduce HoloCosm to preserve score and ranking fidelity across metrics. It achieves the best overall fidelity across 15 benchmark configurations and 85 metrics at 80\% compression while remaining competitive in single-metric evaluation under the same compression setting.

\end{itemize}

\section{Related Work}
\label{sec:related-work}

\paragraph{Benchmark compression.}
Benchmark compression reduces evaluation cost by estimating full-benchmark
performance from fewer items.
AnchorPoints, mRMR, and ZipBench select
representative items using response correlation, mutual-information-based
feature selection, and synthesized-response representations, respectively
\citep{vivek2024anchor,bowyer2026efficient,
huang2026zipbenchlowcostframeworkcompressing}.
tinyBenchmarks and bias-bounded item selection focus on accurate score
estimation through IRT-based selection and bias-bounded submodular selection
\citep{polo2024tinybenchmarks,zhuang2025efficient}. More specialized benchmark
compression includes MiniLongBench, which learns sample representations and
clusters them to prune redundant long-context samples while preserving model
rankings \citep{huang2025minilongbench}. SubLIME uses predicted rank correlation
to guide subset selection, while EssenceBench combines redundancy-aware
filtering with fitness-based subset search
\citep{saranathan2025sublime,wang2026rethinking}. Despite these differences,
existing benchmark-compression methods primarily select items for one
evaluation metric at a time. Tab.~\ref{tab:related_work_comparison} summarizes
this distinction. Our work instead uses multiple evaluation metrics for item
selection within the same benchmark while preserving both scores and model
rankings.

\begin{table}[ht]
\centering
\caption{\textbf{Benchmark-compression methods by selection basis, objective, and metric setting.}}
\label{tab:related_work_comparison}
\small
\begin{tabular}{@{}lllc@{}}
\toprule
\textbf{Method} & \textbf{Selection basis} & \textbf{Objective} & \textbf{Metric setting} \\
\midrule
AnchorPoints
& Response correlation
& Coverage
& Single \\

mRMR
& Mutual information
& Score reconstruction
& Single \\

tinyBenchmarks
& IRT embeddings
& Score estimation
& Single \\

SubLIME
& Predicted rank correlation
& Ranking preservation
& Single \\

EssenceBench
& Redundancy + subset fitness
& Score \& rank fidelity
& Single \\

MiniLongBench
& Learned sample representations
& Ranking preservation
& Single \\

ZipBench
& Synthesized-response representations
& Score \& rank fidelity
& Single \\
\midrule
\textbf{HoloCosm}
& Cross-metric fidelity
& Score \& rank fidelity
& \textbf{Single / Multi} \\
\bottomrule
\end{tabular}
\end{table}

\paragraph{Efficient performance estimation.}
Related work considers multiple benchmarks or exploits cross-metric structure
for more efficient performance estimation
\citep{kipnis2025metabench,liao2025toward}.
These settings do not directly study multi-metric item selection across multiple
evaluation metrics within the same benchmark, which is the focus of our work.

\paragraph{Evaluation reliability.}
Prior work studies the reliability of reduced evaluations and resulting
benchmark rankings
\citep{perlitz2024efficient,yauney2026reliable,neuhof2026quantifying}.
These studies mainly consider reliability under a fixed evaluation criterion.
Our work instead asks whether items selected to represent one metric remain
reliable for other metrics, evaluating both score reconstruction and ranking
preservation consistently across evaluation metrics.

\section{Methodology}
\label{sec:method}

HoloCosm formulates multi-metric benchmark compression as
fidelity-constrained reconstruction. Each candidate set is characterized by
reconstruction accuracy and pairwise-margin distortion across source models
and metrics. Size-matched reference sets calibrate a
ranking-risk threshold. A feasibility-first exchange search seeks
$B$ items with low multi-metric reconstruction error under this constraint,
after which source-assisted estimation estimates full-benchmark target scores.

\paragraph{Preliminaries.}
Let $\mathcal M=[M]$ denote the source-model index set, with source models
$\{f_m\}_{m\in\mathcal M}$, and let
$\mathcal D=\{x_1,\ldots,x_N\}$ denote the benchmark items. Item-level scores for all source models and items under $K$ evaluation metrics form a tensor
$X\in\mathbb R^{M\times N\times K}$, where $X_{mnk}$ is the score of model
$f_m$ on item $x_n$ under metric $k$. For metrics obtained by averaging
item-level scores, the full-benchmark score and its uncorrected subset estimate are
\begin{equation}
\mu_{mk}=\frac1N\sum\nolimits_{n=1}^{N}X_{mnk},
\qquad
\widehat\mu_{mk}(S)=\frac1{|S|}\sum\nolimits_{n\in S}X_{mnk},
\label{eq:mm-full-subset-scores}
\end{equation}
where $n\in[N]$ indexes benchmark items and $S\subseteq[N]$ indexes
the selected items. Collecting these scores gives the full evaluation
matrix $\boldsymbol\mu\in\mathbb R^{M\times K}$ and its subset estimate
$\widehat{\boldsymbol\mu}(S)$. Given a selection size $B<N$, our goal is
to choose a single subset $S$ with $|S|=B$ such that
$\widehat\mu_{mk}(S)\approx\mu_{mk}$ across source models and metrics.
The same selected items are used to estimate every metric. Item selection
uses only source-model responses; held-out target models are evaluated
after $S$ is fixed.

\subsection{Multi-Metric Fidelity Profile}

For every size-$B$ candidate selection $S$, HoloCosm characterizes its
fidelity by reconstruction risk $R_{\mathrm{mean}}(S)$ and ranking risk
$R_{\mathrm{rank}}(S)$. The primary term $R_{\mathrm{mean}}$ averages
normalized reconstruction errors across source models and metrics.
Normalization by the dispersion of full-benchmark source scores makes errors
across metric scales comparable, while retaining every metric in each
candidate assessment. Concretely, for a source pool $\mathcal A$ of $q$
models, let $s_k(\mathcal A)$ be the standard deviation of the full-benchmark
scores for metric $k$ (set to one when numerically zero). We define
\begin{equation}
a_k(S;\mathcal A)
=\frac{1}{q\,s_k(\mathcal A)}
\sum_{m\in\mathcal A}
|\widehat\mu_{mk}(S)-\mu_{mk}|,
\qquad
r_{\mathrm{mean}}(S;\mathcal A)
=\frac{1}{K}\sum_{k=1}^{K}a_k(S;\mathcal A).
\label{eq:main-mean-risk}
\end{equation}

Reconstruction accuracy cannot guarantee faithful model comparisons.
For each metric, $R_{\mathrm{rank}}$ measures reductions in signed pairwise
margins. Let $d_{abk}=\mu_{ak}-\mu_{bk}$,
$\widehat d_{abk}(S)=\widehat\mu_{ak}(S)-\widehat\mu_{bk}(S)$, and let
$p_k(\mathcal A)$ be the median nonzero absolute margin for metric $k$.
Let $\mathcal Q(\mathcal A)$ denote the pooled set of valid
metric-indexed model pairs, given by
$\{(k,a,b):k\in[K],\ (a,b)\in\mathcal P_k(\mathcal A)\}$.
We define the aggregate ranking risk as the mean normalized margin shortfall
over this set:

\begin{equation}
r_{\mathrm{rank}}(S;\mathcal A)
=|\mathcal Q(\mathcal A)|^{-1}
\sum\nolimits_{(k,a,b)\in\mathcal Q(\mathcal A)}
p_k(\mathcal A)^{-1}
\bigl[|d_{abk}|-\operatorname{sign}(d_{abk})
\widehat d_{abk}(S)\bigr]_+.
\label{eq:main-rank-risk}
\end{equation}

This loss penalizes both weakened separation and reversed orderings. The final
risks average these quantities over leave-one-source-out pools,
$R_j(S)=M^{-1}\sum_{m=1}^{M}r_j(S;\mathcal M\setminus\{m\})$ for
$j\in\{\mathrm{mean},\mathrm{rank}\}$, with normalization recomputed in each
pool. Appendix~\ref{app:joint-jackknife} gives the zero-scale conventions,
valid-pair construction, source-pool averaging, and full definitions.

\subsection{Reference-Calibrated Constrained Selection}

HoloCosm minimizes multi-metric reconstruction error subject to a calibrated
constraint on aggregate pairwise-ranking risk. Using an
$\epsilon$-constraint formulation \citep{mesquita2023new}, it solves the
following constrained subset-selection problem:
\begin{equation}
\min_{S\subseteq[N]} R_{\mathrm{mean}}(S)
\quad\text{subject to}\quad
|S|=B,\quad R_{\mathrm{rank}}(S)\leq\epsilon_{\mathrm{rank}}.
\label{eq:concrete-multi-metric-objective}
\end{equation}
Here, reconstruction is the primary objective, while ranking risk is a
feasibility constraint. We draw 128 uniform size-$B$ subsets $\mathcal R$
and set $\epsilon_{\mathrm{rank}}=Q_{0.25}(\{R_{\mathrm{rank}}(S):S\in\mathcal R\})$,
the 25th percentile of their ranking risks. Thus, lower reconstruction
error cannot compensate for risk above this threshold.

\subsection{Feasibility-First Exchange Search}

All candidate selections are ordered by one lexicographic rule. Among
infeasible candidates, smaller normalized constraint violation takes priority.
Every feasible candidate is preferred to every infeasible candidate; within the
feasible region, lower $R_{\mathrm{mean}}$ determines the ordering.

Starting from the best calibrated candidate, HoloCosm repeatedly evaluates
one-for-one exchanges of a selected and an unselected item. An exchange is
accepted only when it improves the lexicographic criterion. Every update
preserves $|S|=B$. 
The initial subset is feasible; accepted exchanges preserve feasibility and reduce reconstruction error.

\subsection{Source-Assisted Score Estimation}
For each target model, we correct its selected-item mean using
source-average scores as auxiliary predictions, following
prediction-powered mean estimation \citep{angelopoulos2023prediction}.
The resulting full-benchmark score estimate is
\begin{equation}
\widetilde\mu_{gk}(S)
=\widehat\mu_{gk}(S)
+N^{-1}\sum\nolimits_{n=1}^{N}z_{nk}
-B^{-1}\sum\nolimits_{n\in S}z_{nk},
\quad
z_{nk}=\frac1M\sum_{m=1}^{M}X_{mnk}.
\label{eq:source-difference-estimator}
\end{equation}
Here, $\widehat\mu_{gk}(S)=B^{-1}\sum_{n\in S}Y_{gnk}$ is the observed
subset mean of target model $g$, where $Y_{gnk}$ is its score on item $n$
under metric $k$. The two sums over $z_{nk}$ are the full-benchmark and
selected-item means of the source-average scores, respectively. Their
difference is shared by all target models for a given metric, so the
correction preserves pairwise score differences and within-metric rankings.
Ranking fidelity is determined by item selection, while the correction
improves full-score estimation without requiring target responses outside
$S$. Algorithm~\ref{alg:constrained-selection} summarizes the procedure.

\begin{algorithm}[H]
\caption{HoloCosm multi-metric subset selection and score estimation}
\label{alg:constrained-selection}
\definecolor{algorithmstage}{RGB}{55,88,112}
\begin{algorithmic}[1]
\REQUIRE Source scores $X$, target models $\mathcal T$, subset size $B$
\ENSURE Selected subset $S$ and corrected target score estimates
\STATE \textcolor{algorithmstage}{\textbf{Calibration:}} Draw a bank $\mathcal R$ of 128 uniform size-$B$ subsets.
\STATE Evaluate $R_{\mathrm{mean}}$ and $R_{\mathrm{rank}}$ for each $S\in\mathcal R$.
\STATE $\epsilon_{\mathrm{rank}}\gets Q_{0.25}(\{R_{\mathrm{rank}}(S):S\in\mathcal R\})$.
\STATE \textcolor{algorithmstage}{\textbf{Selection:}} Initialize $S$ with the best reference subset under the feasibility-first rule.
\FOR{at most 80 refinement rounds}
    \STATE Sample up to 64 selected--unselected item swaps and evaluate both risks for each proposal.
    \STATE Let $S'$ be the best proposal under the feasibility-first rule; \textbf{break} if it does not improve $S$.
    \STATE Accept the improving proposal and update the current subset: $S\gets S'$.
\ENDFOR
\STATE \textcolor{algorithmstage}{\textbf{Estimation:}} Evaluate each target model on $S$.
\STATE Apply the source-assisted correction in Eq.~\ref{eq:source-difference-estimator}.
\end{algorithmic}
\end{algorithm}
\FloatBarrier

\section{Experiments}
\label{sec:evaluation}

\paragraph{Benchmarks and splits.}
We evaluate on 15 benchmark configurations with 85 official
component metrics, as summarized in Tab.~\ref{tab:benchmark_overview}.
They span six LLM, five multimodal-LLM, three VLA, and one agent
configuration, with 12--59 models, 78--8,400 items, and 2--14 metrics each.
Most benchmark matrices are reconstructed from public item-level records;
the remainder use official evaluation pipelines. 
For each configuration, we first compute each model’s full-benchmark mean on every
metric. We orient the metrics so that higher values indicate better performance, 
min–max normalize each metric across the candidate models, 
and take their equally weighted mean as each model’s score.
After ranking models by this score, we select six evenly spaced sources from lowest to highest, using the rest as targets.
All methods use the same fixed source–target split.

\begin{table*}[t]
\caption{
\textbf{Benchmark configurations used in the cross-metric transfer analysis.}
The 15 configurations contain 85 component metrics across LLM, multimodal,
VLA, and agent evaluations.
}
\label{tab:benchmark_overview}
\centering
\small
\setlength{\tabcolsep}{3.2pt}
\renewcommand{\arraystretch}{0.86}
\resizebox{\textwidth}{!}{%
\begin{tabular}{@{}lllrrr@{}}
\toprule
\textbf{Type}
& \textbf{Benchmark}
& \textbf{Task}
& \textbf{Models}
& \textbf{Items}
& \textbf{Metrics} \\
\midrule
\multirow{6}{*}{\textbf{LLM}}
& ConStory \citep{li2026lost} & Story evaluation & 33 & 1,179 & 5 \\
& LiveIdea-Core3 \citep{ruan2026evaluating} & Ideation & 41 & 869 & 3 \\
& LiveIdea-Full4 \citep{ruan2026evaluating} & Ideation & 41 & 235 & 4 \\
& LongForm \citep{paech2025longform} & Writing & 59 & 108 & 14 \\
& MMDocRAG-Text \citep{dong2026benchmarking} & Retrieval & 51 & 1,955 & 8 \\
& RouterArena \citep{lu2026routerarena} & Routing & 12 & 8,400 & 2 \\
\midrule
\multirow{5}{*}{\textbf{MLLM}}
& MMDocRAG-Multimodal \citep{dong2026benchmarking} & Retrieval & 31 & 1,974 & 8 \\
& SoB-Text \citep{singh2026structured} & Structured output & 35 & 5,000 & 6 \\
& SoB-Image \citep{singh2026structured} & Structured output & 35 & 209 & 5 \\
& SoB-Audio \citep{singh2026structured} & Structured output & 34 & 115 & 5 \\
& Widget2Code \citep{zhang2026widget2code} & Code generation & 13 & 288 & 12 \\
\midrule
\multirow{3}{*}{\textbf{VLA}}
& Bench2Drive \citep{jia2024bench2drive} & Driving & 32 & 220 & 4 \\
& Fail2Drive-ID \citep{gerstenecker2026fail2drive} & Driving & 33 & 82 & 3 \\
& Fail2Drive-OOD \citep{gerstenecker2026fail2drive} & Driving & 33 & 78 & 3 \\
\midrule
\textbf{Agent}
& WorkBench \citep{styles2026workbench} & Tool use & 24 & 690 & 3 \\
\bottomrule
\end{tabular}%
}

\end{table*}

\paragraph{Baselines.}
We compare HoloCosm with Random, tinyBenchmarks
\citep{polo2024tinybenchmarks}, ZipBench
\citep{huang2026zipbenchlowcostframeworkcompressing},
mRMR with Ridge reconstruction \citep{bowyer2026efficient},
AnchorPoints \citep{vivek2024anchor}, and EssenceBench
\citep{wang2026rethinking}.
Random performs uniform random sampling without replacement.
We adapt the baselines to the common six-source-model protocol;
implementation details are provided in Appendix~\ref{app:external-baselines}.

\begin{figure*}[ht]
\centering
\includegraphics[
  width=\linewidth,
  trim=0pt 0pt 0pt 0pt,
  clip
]{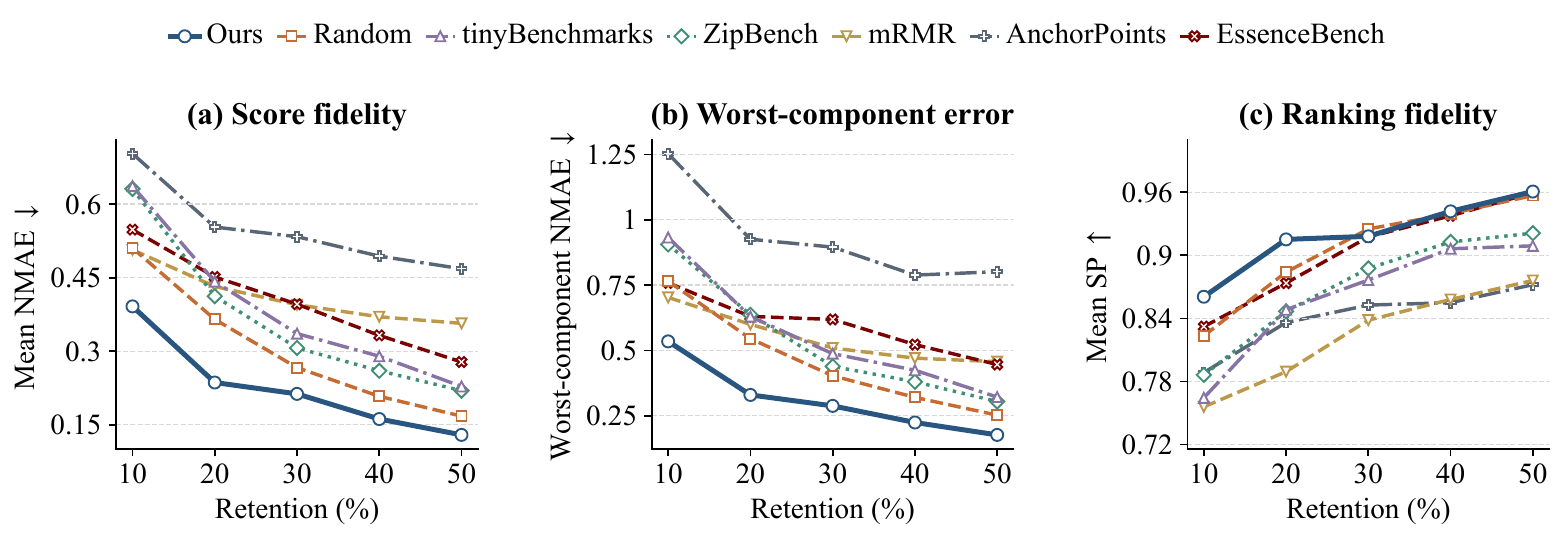}
\setlength{\abovecaptionskip}{4pt}
\setlength{\belowcaptionskip}{4pt}
\caption{\textbf{Multi-metric fidelity across retention ratios.}
Mean NMAE, worst-component NMAE, and mean SP are computed across evaluation
metrics within each benchmark, averaged over runs, then equally across 15 benchmarks.
For each single-target baseline, the target with lowest run-averaged mean
NMAE is selected at each benchmark and retention level and used for all panels.
} 
\label{fig:main_comparison}
\end{figure*}

\paragraph{Comparison protocol.}
We test retention ratios of 10\%, 20\%, 30\%, 40\%, and 50\%,
with 20\% as the main setting. For $N$ items at retention ratio $\rho$,
we set the selection size to the nearest integer to $\rho N$, with at least
one item. At each ratio, every method selects the same number of unique items.
Each single-target baseline selects items using each component metric in turn
and evaluates the resulting selection on all components. HoloCosm selects items across all components, whereas Random uniformly
samples items and evaluates all components on those items.

\paragraph{Evaluation metrics.}
We measure score reconstruction using normalized mean absolute error (NMAE)
and ranking preservation using Spearman rank correlation (SP) across target
models. NMAE normalizes reconstruction error by target-score dispersion,
making errors comparable across metrics with different scales.
For component $k$ of benchmark $b$, we define NMAE as
\begin{equation}
\operatorname{NMAE}_{b,k}
=
\frac{
|\mathcal T_b|^{-1}
\sum_{t\in\mathcal T_b}
|\widehat y_{t,k}-y_{t,k}|
}{
\sigma_{b,k}
},
\qquad
\sigma_{b,k}
=
\operatorname{Std}_{t\in\mathcal T_b}(y_{t,k}).
\label{eq:eval-nmae}
\end{equation}
Here, $\mathcal T_b$ denotes the target models, and $y_{t,k}$ and
$\widehat y_{t,k}$ denote the full-benchmark and estimated scores,
respectively. $\sigma_{b,k}$ is the population standard deviation across
target models. SP measures agreement between rankings induced by the full
and estimated scores. For each run, we report mean and worst-component
values, using maximum NMAE or minimum SP. Results are averaged over runs
and macro-averaged across benchmarks. Lower NMAE and higher SP indicate
better fidelity.

\subsection{Main results}
\label{sec:results}
\label{sec:main_results}

For each single-metric baseline, we report the selection target with the
lowest mean NMAE for each benchmark and retention level, and use the same
target for all measures.

{\textbf{Results at 20\% retention.}}
Tab.~\ref{tab:retention20_cross_metric_nmae} reports the main results at
20\% retention. HoloCosm achieves the best aggregate fidelity, while
Random is the strongest baseline overall. Compared with Random, HoloCosm
reduces mean and worst-component NMAE by 35.3\% and 39.5\%, respectively,
while also improving ranking fidelity. In contrast, all single-metric baselines
underperform Random in macro-average NMAE and SP, with NMAE ranging from
0.41 to 0.55 and SP from 0.79 to 0.87. This advantage is also broadly consistent across
benchmarks: HoloCosm outperforms Random in mean NMAE on 14 of 15 benchmarks
and in mean SP on 11 of 15.

{\textbf{Across retention levels.}}
To assess performance beyond the 20\% setting,
Fig.~\ref{fig:main_comparison} compares HoloCosm with all baselines from
10\% to 50\% retention. HoloCosm achieves the lowest mean and worst-component
NMAE at every level, while maintaining consistently competitive ranking fidelity
throughout the full tested retention range. Both error measures decrease steadily
and mean SP rises as more items are retained. The margin is clearest under
aggressive compression and persists at higher retention, showing that the gains
are not tied to a single operating point.

\begin{table*}[t]
\setlength{\belowcaptionskip}{4pt}
\centering
\caption{\textbf{Multi-metric score and ranking fidelity at 20\% retention.}
Entries average over $K$ metrics and runs with benchmarks weighted equally; each
baseline's best-NMAE target supplies the measures. W/T/L denotes wins, ties, and
losses against Random. Bold indicates best results.}
\label{tab:retention20_cross_metric_nmae}
\setlength{\tabcolsep}{2.5pt}
\renewcommand{\arraystretch}{1.28}
\resizebox{\linewidth}{!}{%
\begin{tabular}{lccccccccccccccc}
\toprule
& & \multicolumn{4}{c}{All-metric / random selection}& \multicolumn{10}{c}{Single-metric selection (best across metrics)} \\
\cmidrule(lr){3-6}\cmidrule(lr){7-16}

& & \multicolumn{2}{c}{Ours} & \multicolumn{2}{c}{Random} & \multicolumn{2}{c}{tinyBenchmarks} & \multicolumn{2}{c}{ZipBench} & \multicolumn{2}{c}{mRMR} & \multicolumn{2}{c}{AnchorPoints} & \multicolumn{2}{c}{EssenceBench} \\
\cmidrule(lr){3-4}\cmidrule(lr){5-6}\cmidrule(lr){7-8}\cmidrule(lr){9-10}\cmidrule(lr){11-12}\cmidrule(lr){13-14}\cmidrule(lr){15-16}
Benchmark & $K$ & NMAE $\downarrow$ & SP $\uparrow$ & NMAE $\downarrow$ & SP $\uparrow$ & NMAE $\downarrow$ & SP $\uparrow$ & NMAE $\downarrow$ & SP $\uparrow$ & NMAE $\downarrow$ & SP $\uparrow$ & NMAE $\downarrow$ & SP $\uparrow$ & NMAE $\downarrow$ & SP $\uparrow$ \\
\midrule
ConStory & 5 & 0.30 & 0.88 & \textbf{0.29} & 0.88 & 0.43 & 0.84 & 0.45 & 0.81 & 0.67 & 0.69 & 1.24 & 0.52 & 0.44 & \textbf{0.88} \\
Bench2Drive & 4 & \textbf{0.27} & 0.84 & 0.39 & 0.84 & 0.46 & 0.78 & 0.53 & 0.76 & 0.37 & 0.83 & 0.59 & 0.79 & 0.34 & \textbf{0.85} \\
Fail2Drive-ID & 3 & \textbf{0.28} & \textbf{0.80} & 0.58 & 0.59 & 0.74 & 0.79 & 0.51 & 0.57 & 0.37 & 0.49 & 0.28 & 0.71 & 1.00 & 0.51 \\
Fail2Drive-OOD & 3 & \textbf{0.39} & \textbf{0.86} & 0.79 & 0.79 & 0.68 & 0.69 & 0.73 & 0.65 & 0.50 & 0.51 & 0.61 & 0.80 & 0.58 & 0.76 \\
LiveIdea-Core3 & 3 & 0.10 & 0.98 & 0.11 & 0.98 & 0.16 & 0.96 & 0.16 & 0.96 & 0.33 & 0.97 & 0.12 & 0.98 & \textbf{0.10} & \textbf{0.98} \\
LiveIdea-Full4 & 4 & \textbf{0.22} & \textbf{0.95} & 0.29 & 0.94 & 0.39 & 0.87 & 0.40 & 0.88 & 0.42 & 0.89 & 0.27 & 0.94 & 0.25 & 0.94 \\
RouterArena & 2 & \textbf{0.09} & 0.97 & 0.15 & \textbf{0.99} & 0.28 & 0.91 & 0.25 & 0.96 & 1.02 & 0.86 & 0.28 & 0.97 & 0.42 & 0.93 \\
LongForm & 14 & \textbf{0.12} & \textbf{0.98} & 0.14 & 0.98 & 0.17 & 0.97 & 0.17 & 0.97 & 0.13 & 0.98 & 0.15 & 0.98 & 0.32 & 0.96 \\
SoB-Audio & 5 & 0.58 & \textbf{0.83} & 0.70 & 0.78 & 0.99 & 0.65 & 0.78 & 0.75 & \textbf{0.56} & 0.75 & 0.72 & 0.80 & 0.58 & 0.74 \\
SoB-Image & 5 & \textbf{0.51} & \textbf{0.77} & 1.17 & 0.65 & 1.03 & 0.62 & 0.92 & 0.71 & 0.56 & 0.66 & 0.87 & 0.71 & 1.41 & 0.70 \\
SoB-Text & 6 & \textbf{0.16} & \textbf{0.96} & 0.21 & 0.96 & 0.39 & 0.86 & 0.39 & 0.87 & 0.51 & 0.88 & 1.80 & 0.67 & 0.46 & 0.96 \\
MMDocRAG-MM & 8 & \textbf{0.06} & \textbf{0.99} & 0.07 & 0.99 & 0.14 & 0.98 & 0.11 & 0.98 & 0.10 & 0.99 & 0.09 & 0.99 & 0.08 & 0.99 \\
MMDocRAG-Text & 8 & \textbf{0.10} & 0.98 & 0.11 & 0.98 & 0.14 & 0.97 & 0.17 & 0.96 & 0.23 & 0.97 & 0.13 & 0.97 & 0.11 & \textbf{0.98} \\
Widget2Code & 12 & \textbf{0.22} & \textbf{0.94} & 0.30 & 0.93 & 0.42 & 0.88 & 0.38 & 0.90 & 0.24 & 0.92 & 0.33 & 0.86 & 0.30 & 0.94 \\
WorkBench & 3 & \textbf{0.14} & \textbf{0.98} & 0.17 & 0.97 & 0.22 & 0.95 & 0.22 & 0.97 & 0.48 & 0.46 & 0.83 & 0.84 & 0.38 & 0.96 \\
\midrule
Macro-average & -- & \textbf{0.24} & \textbf{0.92} & 0.37 & 0.88 & 0.44 & 0.85 & 0.41 & 0.85 & 0.43 & 0.79 & 0.55 & 0.84 & 0.45 & 0.87 \\
W/T/L vs. Random & -- & \textbf{14/0/1} & \textbf{11/0/4} & -- & -- & 2/0/13 & 1/0/14 & 3/0/12 & 1/0/14 & 7/0/8 & 1/0/14 & 4/0/11 & 5/0/10 & 5/0/10 & 8/0/7 \\
\bottomrule
\end{tabular}%
}
\end{table*}

\begin{table*}[t]
\setlength{\abovecaptionskip}{2pt}
\setlength{\belowcaptionskip}{4pt}
\caption{\textbf{Cross-metric fidelity on Bench2Drive at 20\% retention.}
Underlining marks selection-metric results; Mean averages NMAE, and blue-gray marks each baseline\textquotesingle s lowest-mean target.}
\label{tab:drive_cross_metric_compact}
\centering
\scriptsize
\setlength{\tabcolsep}{3.2pt}
\renewcommand{\arraystretch}{1.02}
\begin{tabular}{@{}llccccccccc@{}}
\toprule
\textbf{Method} & \textbf{Selection metric}
& \multicolumn{5}{c}{\textbf{NMAE $\downarrow$}}
& \multicolumn{4}{c}{\textbf{SP $\uparrow$}} \\
\cmidrule(lr){3-7}\cmidrule(lr){8-11}
 &  & \textbf{Efficiency} & \textbf{Score} & \textbf{Smoothness} & \textbf{Success} & \textbf{Mean} & \textbf{Efficiency} & \textbf{Score} & \textbf{Smoothness} & \textbf{Success} \\
\midrule
\rowcolor{black!5}
Ours & All metrics & 0.053 & 0.435 & 0.159 & 0.448 & 0.274 & 0.782 & 0.845 & 0.911 & 0.835 \\
\midrule
\multirow{4}{*}{ZipBench}
 & Efficiency & \underline{0.111} & 0.872 & 0.343 & 0.818 & 0.536 & \underline{0.758} & 0.748 & 0.862 & 0.787 \\
 & \bestcell{Score} & \bestcell{0.100} & \bestcell{\underline{0.831}} & \bestcell{0.338} & \bestcell{0.860} & \bestcell{0.532} & \bestcell{0.668} & \bestcell{\underline{0.750}} & \bestcell{0.869} & \bestcell{0.747} \\
 & Smoothness & 0.174 & 0.944 & \underline{0.299} & 0.973 & 0.598 & 0.847 & 0.783 & \underline{0.845} & 0.825 \\
 & Success & 0.229 & 0.807 & 0.315 & \underline{0.821} & 0.543 & 0.689 & 0.815 & 0.872 & \underline{0.726} \\
\bottomrule
\end{tabular}

\end{table*}

\FloatBarrier
\textbf{Cross-metric transfer.}
Tab.~\ref{tab:retention20_cross_metric_nmae} reports each single-metric
baseline's results using the selection metric with the lowest mean NMAE.
Tab.~\ref{tab:drive_cross_metric_compact} details ZipBench on Bench2Drive.
ZipBench achieves its lowest mean NMAE with Score as the compression target.
Switching to Smoothness lowers Smoothness error but increases Efficiency error.
HoloCosm jointly lowers all four component errors,
avoiding this target-dependent trade-off.
Complete results are provided in
Appendix~\ref{app:small_metric_cross_results}.

\subsection{Ablation Study}
\label{sec:ablation}
\label{sec:joint_objective_ablation}

\begin{table}[H]
\caption{\textbf{Selection ablation at 20\% retention.}
All variants use the same source-assisted estimator. Results are three-seed
means with benchmarks weighted equally.}
\label{tab:additional_ablation}
\small
\setlength{\tabcolsep}{3.5pt}
\renewcommand{\arraystretch}{1.08}
\resizebox{\linewidth}{!}{%
\begin{tabular}{@{}llrrr@{}}
\toprule
Variant
& Selection rule
& Mean NMAE $\downarrow$
& Worst NMAE $\downarrow$
& Mean SP $\uparrow$ \\
\midrule
Random + correction
& Uniform sampling
& 0.2812 & 0.3828 & 0.8840 \\

Multi-metric reconstruction
& Minimize $R_{\mathrm{mean}}$
& 0.2404 & 0.3390 & 0.9077 \\

HoloCosm
& Minimize $R_{\mathrm{mean}}$ s.t. $R_{\mathrm{rank}}\leq\epsilon_{\mathrm{rank}}$
& \textbf{0.2362}
& \textbf{0.3299}
& \textbf{0.9151} \\
\bottomrule
\end{tabular}%
}
\end{table}

As shown in Tab.~\ref{tab:additional_ablation}, we isolate the contribution
of each selection component at 20\% retention across 15 benchmarks, while
keeping the source-assisted estimator unchanged. We compare uniform random
sampling with source-assisted correction, multi-metric reconstruction without
explicit ranking protection, and the full HoloCosm selection with
ranking-aware constraints.

Multi-metric reconstruction improves score and ranking fidelity over
random sampling with correction, reducing mean NMAE from 0.2812 to 0.2404
and raising mean SP from 0.884 to 0.9077. Adding ranking protection yields
gains, reaching 0.2362 mean NMAE, 0.3299 worst-component NMAE, and 0.9151
mean SP. This shows that ranking-aware selection improves not only ordering
but also reconstruction fidelity, including the hardest-to-preserve
components. Since source-assisted correction preserves model ordering, the
SP gains reflect the selection.

\subsection{Additional Analyses}

\par\addvspace{0.6\baselineskip}
\noindent\begingroup%
\setlength{\fboxsep}{7pt}\setlength{\fboxrule}{0.4pt}%
\definecolor{questionbackground}{HTML}{EEF3F8}%
\fcolorbox{black}{questionbackground}{%
\parbox{\dimexpr\linewidth-2\fboxsep-2\fboxrule\relax}{%
\textcolor[HTML]{174A7E}{\textbf{(1)}}\enspace\textbf{Can Simple Multi-Metric Strategies Preserve Benchmark Scores and Rankings?}}}%
\endgroup\par\nobreak\vspace{3pt}
\begingroup
\setlength{\intextsep}{0pt}
\setlength{\abovecaptionskip}{0pt}
\setlength{\belowcaptionskip}{3pt}
\begin{table}[H]
\centering
\caption{Simple multi-metric compression strategies at
20\% retention. Results are averaged equally over 15 configurations;
mean and worst-metric errors are computed per configuration.}
\label{tab:simple_multimetric}
\small
\begin{tabular*}{\linewidth}{@{\extracolsep{\fill}}llccc@{}}
\toprule
Method & Strategy & Mean NMAE $\downarrow$ & Worst NMAE $\downarrow$ & Mean SP $\uparrow$ \\
\midrule
Random & --- & 0.3650 & 0.5450 & 0.8840 \\
\midrule
\multirow{3}{*}{AnchorPoints}
& Single metric & 0.5533 & 0.9256 & 0.8361 \\
& AverageMetric & 0.6106 & 1.1302 & 0.8497 \\
& Vote & 0.6288 & 0.9719 & 0.9038 \\
\midrule
K-means & Multi-metric responses & 0.3242 & 0.4288 & 0.8842 \\
\midrule
\multirow{2}{*}{Ours}
& AverageMetric & 0.2822 & 0.3896 & 0.8946 \\
& All metrics & \textbf{0.2362} & \textbf{0.3299} & \textbf{0.9151} \\
\bottomrule
\end{tabular*}
\end{table}
\endgroup

\noindent
To test whether simple multi-metric strategies can preserve benchmark scores
and rankings, we compare three strategies on the original metrics at 20\%
retention.

\noindent
\textbf{AverageMetric:} We direction-align all metrics, normalize each using
source-derived min--max statistics, and average them equally into a scalar
selection target. For AnchorPoints and Ours, this scalar serves as the sole selection
target, while their original score estimators remain unchanged.

\noindent
\textbf{Vote:} We run AnchorPoints for each metric to select $B$ items, giving candidates one vote per metric.
We select by vote count; boundary ties favor the least-covered
metrics, followed by tie-breaking. This yields $B$ distinct items, whose
unweighted means estimate target metrics.

\noindent
\textbf{K-means:} We concatenate each item's source responses after scaling
metrics by the standard deviation of source full-benchmark means. Lloyd K-means
forms $B$ clusters and retains the item nearest each center. Target scores are
cluster-size-weighted averages of the representatives.

Tab.~\ref{tab:simple_multimetric} shows uneven gains.
Vote improves ranking correlation but increases score error,
and random selection remains a competitive baseline.
Ours achieves lower error and higher ranking agreement.
Compared with K-means, all-metric selection reduces mean and worst NMAE by
27.1\% and 23.1\%, respectively, while increasing mean SP from 0.8842 to
0.9151. Thus, directly optimizing metric-specific response structure is more
effective than generic response-space clustering.

\par\addvspace{0.6\baselineskip}
\noindent\begingroup%
\setlength{\fboxsep}{7pt}\setlength{\fboxrule}{0.4pt}%
\definecolor{questionbackground}{HTML}{EEF3F8}%
\fcolorbox{black}{questionbackground}{%
\parbox{\dimexpr\linewidth-2\fboxsep-2\fboxrule\relax}{%
\textcolor[HTML]{174A7E}{\textbf{(2)}}\enspace\textbf{Can HoloCosm Match Strong Baselines on Single-Metric Benchmarks?}}}%
\endgroup\par\nobreak\vspace{3pt}
\begingroup
\setlength{\intextsep}{0pt}
\setlength{\abovecaptionskip}{0pt}
\setlength{\belowcaptionskip}{3pt}
\begin{table}[H]
\centering
\caption{
\textbf{Single-metric evaluation at 20\% retention.}
MAE measures score reconstruction and SP measures ranking preservation.
Results are averaged over runs; Macro-average weights the three benchmarks equally.
Bold marks the best result in each column.
}
\label{tab:single_metric_evaluation}
\small
\setlength{\tabcolsep}{5pt}
\renewcommand{\arraystretch}{1.10}
\resizebox{\textwidth}{!}{%
\begin{tabular}{lcccccccc}
\toprule
& \multicolumn{2}{c}{\textbf{MMStar}}
& \multicolumn{2}{c}{\textbf{SimpleQA}}
& \multicolumn{2}{c}{\textbf{SimpleVQA}}
& \multicolumn{2}{c}{\textbf{Macro-average}} \\
\cmidrule(lr){2-3}
\cmidrule(lr){4-5}
\cmidrule(lr){6-7}
\cmidrule(lr){8-9}
Method
& MAE $\downarrow$ & SP $\uparrow$
& MAE $\downarrow$ & SP $\uparrow$
& MAE $\downarrow$ & SP $\uparrow$
& MAE $\downarrow$ & SP $\uparrow$ \\
\midrule
\rowcolor{black!5}
Ours
& 0.017 & 0.964
& 0.009 & \textbf{1.000}
& \textbf{0.013} & \textbf{1.000}
& \textbf{0.013} & 0.988 \\

Random
& \textbf{0.014} & \textbf{0.988}
& \textbf{0.009} & \textbf{1.000}
& 0.022 & 0.992
& 0.015 & \textbf{0.993} \\

tinyBenchmarks
& 0.032 & 0.952
& 0.022 & 0.988
& 0.018 & 0.984
& 0.024 & 0.975 \\

ZipBench
& 0.018 & 0.964
& 0.018 & 0.988
& 0.031 & 0.937
& 0.023 & 0.963 \\

mRMR
& 0.067 & 0.893
& 0.030 & \textbf{1.000}
& 0.022 & 0.976
& 0.040 & 0.956 \\

AnchorPoints
& 0.125 & 0.810
& 0.119 & 0.881
& 0.156 & 0.262
& 0.134 & 0.651 \\
\bottomrule
\end{tabular}%
}
\end{table}
\endgroup

\noindent
We further examine whether joint multi-metric optimization compromises performance
when only one metric is evaluated on three single-metric benchmarks:
MMStar, SimpleQA, and SimpleVQA
\citep{chen2024we,wei2024measuring,cheng2025simplevqa},
spanning diverse evaluation settings.

As shown in
Tab.~\ref{tab:single_metric_evaluation}, at 20\% retention
HoloCosm achieves the lowest macro-average MAE of 0.013 and a macro-average
SP of 0.988, close to Random's 0.993, while outperforming all existing
compression baselines. These results show that HoloCosm's effectiveness
is not limited to multi-metric evaluation and remains competitive when
only a single metric is available.

\par\addvspace{0.6\baselineskip}
\noindent\begingroup%
\setlength{\fboxsep}{7pt}\setlength{\fboxrule}{0.4pt}%
\definecolor{questionbackground}{HTML}{EEF3F8}%
\fcolorbox{black}{questionbackground}{%
\parbox{\dimexpr\linewidth-2\fboxsep-2\fboxrule\relax}{%
\textcolor[HTML]{174A7E}{\textbf{(3)}}\enspace\textbf{Can Weaker Source Models Guide Benchmark Compression for Stronger Models?}}}%
\endgroup\par
\begin{figure}[!htb]
\centering
\setlength{\abovecaptionskip}{4pt}
\setlength{\belowcaptionskip}{0pt}
\includegraphics[width=\linewidth]{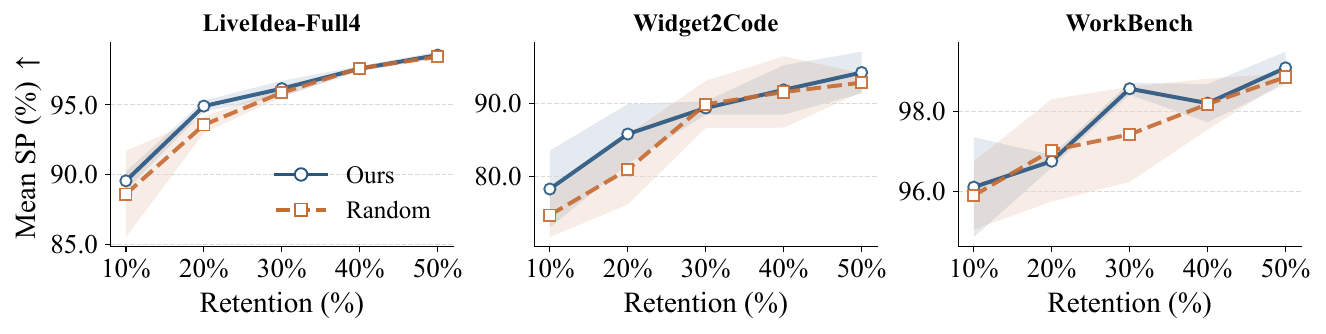}
\caption{\textbf{Weak-to-strong generalization.}
Six weaker models guide selection for disjoint stronger models. Curves show
mean SP $\pm$ standard deviation over three seeds.}
\label{fig:weak-to-strong-generalization}
\end{figure}

\nobreak\vspace{0.35\baselineskip}
\noindent
\label{sec:weak_to_strong}
To evaluate transfer beyond the source pool's capability,
we construct a weak-to-strong setting. We rank models by the
four-metric average on LiveIdea-Full4,
SSIM on Widget2Code, and Correct on WorkBench. Six weaker models guide
selection for disjoint stronger models.
Fig.~\ref{fig:weak-to-strong-generalization} shows that ranking fidelity
improves with retention. HoloCosm's advantage over Random is
most pronounced at low retention on LiveIdea-Full4 and Widget2Code,
while the two methods perform similarly on WorkBench. The curves converge
as retention increases. These results suggest that weaker
source models can guide selection for stronger targets, with gains
varying across benchmarks.

\section{Conclusion}
\label{sec:conclusion}
We uncover a systematic cross-metric failure in benchmark compression:
subsets selected for one metric often perform worse than equal-size random
sampling on others. To address this, HoloCosm minimizes multi-metric
reconstruction error under a ranking-risk constraint. Across 15 benchmark
configurations, HoloCosm improves score and ranking fidelity over Random.
It remains competitive on single-metric benchmarks and transfers from
weaker source models to stronger targets. Benchmark compression for
multi-metric evaluation should preserve fidelity across metrics.

\bibliography{references}
\bibliographystyle{references}

\clearpage
\appendix
\makeatletter
\setlength{\@fptop}{0pt}
\setlength{\@fpsep}{18pt}
\setlength{\@fpbot}{0pt plus 1fil}
\makeatother

\section*{\textcolor{blue!45!black}{Appendix Contents}}
\begin{center}
{\small
\begin{tabularx}{\linewidth}{@{}p{3.2em}Xr@{}}
\appmain{app:method}
\appsub{app:notation}
\appsub{app:joint-jackknife}
\appsub{app:algorithms}
\appsub{app:source-correction}
\appsub{app:error-analysis}
\addlinespace[0.35em]
\appmain{app:experimental-setup}
\appsub{app:external-baselines}
\appsub{app:simple-multimetric-details}
\addlinespace[0.35em]
\appmain{app:analysis-discussion}
\appsub{app:random-scores}
\appsub{app:one-metric-sufficient}
\appsub{app:alternative-strategies}
\appsub{app:source-model-selection}
\appsub{app:additional-ablation}
\appsub{app:evaluation-cost}
\addlinespace[0.35em]
\appmain{app:results}
\appsub{app:all-target-results}
\appsub{app:strategy-results}
\appsub{app:small_metric_cross_results}
\end{tabularx}
}
\end{center}
\clearpage

\section{Methodological Details}
\label{app:method}

\subsection{Problem Formulation and Notation}
\label{app:notation}
\label{app:extended-formulation}
\label{app:abstract-formulation}

Let \(X_{mnk}\) denote the score of source model \(m\) on item \(n\) under metric \(k\), and let \(Y_{gnk}\) denote the corresponding score of target model \(g\). The benchmark contains \(N\ge2\) items and \(K\) metrics. We select \(S\subseteq[N]\) with \(|S|=B\), where \(1\le B<N\). For any model, its full-benchmark score and subset estimate are the corresponding arithmetic means over all items and over \(S\), respectively.

For a target model, define
\[
\mu_{gk}=\frac1N\sum_{n=1}^{N}Y_{gnk},
\qquad
\widehat\mu_{gk}(S)=\frac1B\sum_{n\in S}Y_{gnk},
\]
and its signed estimation error as \(\delta_{gk}(S)=\widehat\mu_{gk}(S)-\mu_{gk}\). The same definitions apply to source models by replacing \(Y\) with \(X\). Recall that \(\mathcal M=[M]\) is the source-model index set.

Multi-metric compression jointly uses all $K$ metrics to select a subset $S$ of $B$ items. The following sections define the source-based reconstruction objective and ranking-risk constraint used to select this subset.

\subsection{Multi-Metric Selection Risks}
\label{app:joint-jackknife}

This section defines the multi-metric reconstruction objective and the pairwise ranking risk used by the selector. All quantities are computed from source scores. Let \(\mathcal A\subseteq\mathcal M\) be a source-model pool containing \(q\) models.

\subsubsection{Normalized Mean Error}
\label{app:mean-risk}

For each metric, normalize errors by the population standard deviation of full-benchmark scores across models in \(\mathcal A\):
\[
\sigma_k(\mathcal A)
=
\sqrt{
\frac1q
\sum_{m\in\mathcal A}
\left(\mu_{mk}-\overline\mu_{\mathcal A,k}\right)^2
},
\qquad
\overline\mu_{\mathcal A,k}
=
\frac1q\sum_{m\in\mathcal A}\mu_{mk}.
\]
Use \(s_k(\mathcal A)=\sigma_k(\mathcal A)\) when \(\sigma_k(\mathcal A)>\eta\), and \(s_k(\mathcal A)=1\) otherwise, where \(\eta\) is machine precision. Define the normalized model error and the reconstruction error of each metric as
\[
e_{mk}(S;\mathcal A)
=
\frac{|\widehat\mu_{mk}(S)-\mu_{mk}|}{s_k(\mathcal A)},
\qquad
a_k(S;\mathcal A)=\frac1q\sum_{m\in\mathcal A}e_{mk}(S;\mathcal A).
\]
The multi-metric mean risk is
\[
r_{\mathrm{mean}}(S;\mathcal A)
=\frac1K\sum_{k=1}^{K}a_k(S;\mathcal A)
=\frac1{qK}\sum_{m\in\mathcal A}\sum_{k=1}^{K}e_{mk}(S;\mathcal A).
\]

Normalization makes the metric errors comparable, and the outer average gives each metric equal weight. Crucially, all metric terms are evaluated together for each candidate \(S\): a proposed exchange is assessed through its combined effect across metrics. A single-metric reconstruction criterion contains only one component \(a_k\), while the multi-metric criterion includes all \(K\) components. Source-pool averaging below preserves this decomposition.

\subsubsection{Pairwise Margin Risk}
\label{app:rank-risk}

For each model pair and metric, define \(d_{abk}=\mu_{ak}-\mu_{bk}\) and \(\widehat d_{abk}(S)=\widehat\mu_{ak}(S)-\widehat\mu_{bk}(S)\). Let
\[
\mathcal P_k(\mathcal A)
=
\{(a,b):a,b\in\mathcal A,\ a<b,\ |d_{abk}|>\eta\}.
\]
For nonempty \(\mathcal P_k(\mathcal A)\), use the scale \(p_k(\mathcal A)=\operatorname{median}_{(a,b)\in\mathcal P_k(\mathcal A)}|d_{abk}|\); set \(p_k(\mathcal A)=1\) when the pair set is empty. The ranking risk is
\[
r_{\mathrm{rank}}(S;\mathcal A)
=
\frac{
\displaystyle
\sum_{k=1}^{K}
\sum_{(a,b)\in\mathcal P_k(\mathcal A)}
\frac{
\left[
|d_{abk}|
-
\operatorname{sign}(d_{abk})\widehat d_{abk}(S)
\right]_+
}{p_k(\mathcal A)}
}{
\displaystyle\sum_{k=1}^{K}|\mathcal P_k(\mathcal A)|
},
\]
where \([u]_+=\max(u,0)\). If there are no valid pairs, the risk is zero.

The loss penalizes a reduction in the correctly oriented score gap and increases further when the ordering reverses. Valid pair–metric entries are pooled into one average.

\subsubsection{Source-Pool Risk Averaging}
\label{app:source-pool-averaging}

HoloCosm evaluates each candidate selection across overlapping source-model pools. With \(M\) source models, define \(\mathcal A_{-m}=\mathcal M\setminus\{m\}\) and
\begin{equation}
R_j(S)
=
\frac1M\sum_{m=1}^{M}r_j(S;\mathcal A_{-m}),
\qquad
j\in\{\mathrm{mean},\mathrm{rank}\}.
\label{eq:app-source-pool-risk}
\end{equation}

The same candidate \(S\) is evaluated in every pool. Normalization scales and pairwise losses are recomputed using the models in that pool. With six source models, this produces six five-model risk evaluations.

This averaging incorporates several source-pool compositions into each candidate assessment. In the implementation, it is enabled when at least four source models are available; with three models, risks are evaluated directly on the full source pool.

\subsection{Constraint Calibration and Exchange Search}
\label{app:algorithms}

\paragraph{Constrained selection.}
\label{app:method-derivation}

The $\epsilon$-constraint approach optimizes a primary objective while constraining another criterion \citep{mesquita2023new}. We apply it to multi-metric reconstruction error and pairwise-ranking distortion:
\[
\min_{S:\,|S|=B}R_{\mathrm{mean}}(S)
\quad
\text{subject to}\quad
R_{\mathrm{rank}}(S)\le\epsilon_{\mathrm{rank}}.
\]

The mean objective combines score reconstruction across metrics, and the ranking constraint controls pairwise margin distortion.

\subsubsection{Constraint Calibration}
\label{app:thresholds}

We draw \(L=128\) uniform size-\(B\) candidate selections, sampling without replacement within each selection. The ranking threshold is
\begin{equation}
\epsilon_{\mathrm{rank}}
=
Q_{0.25}\bigl(\{R_{\mathrm{rank}}(S_\ell)\}_{\ell=1}^{L}\bigr),
\label{eq:app-ranking-threshold}
\end{equation}
where \(Q_{0.25}\) uses linear interpolation between ordered observations.

This threshold expresses the desired ranking quality relative to selections of the same size. It remains fixed throughout refinement. The candidate pool also supplies the initial selection.

\subsubsection{Exchange Search and Its Properties}
\label{app:comparison}

Define the relative excess
\[
E(v,b)=
\begin{cases}
\max(0,v/b-1), & b>\eta,\\
0, & b\le\eta\ \text{and}\ v\le\eta,\\
v/\eta, & b\le\eta\ \text{and}\ v>\eta.
\end{cases}
\]
The ranking-constraint violation is \(V(S)=E(R_{\mathrm{rank}}(S),\epsilon_{\mathrm{rank}})\). Candidates are compared lexicographically using
\begin{equation}
\Psi(S)=
\begin{cases}
(0,0,R_{\mathrm{mean}}(S)),&V(S)=0,\\
(1,V(S),R_{\mathrm{mean}}(S)),&V(S)>0.
\end{cases}
\label{eq:app-candidate-comparison}
\end{equation}

A candidate is numerically feasible when \(V(S)=0\), using tolerance \(\eta\) when \(\epsilon_{\mathrm{rank}}\leq\eta\). Feasible candidates are preferred; among feasible candidates, lower mean error is preferred. For infeasible candidates, lower constraint violation takes priority, followed by mean error.

\paragraph{One-for-one refinement.}
\label{app:swap-procedure}

Initialize \(S\) with the candidate having the smallest comparison key. At each round, sample up to \(64\) distinct exchange pairs \((i,j)\in S\times([N]\setminus S)\) and evaluate \(S'=(S\setminus\{i\})\cup\{j\}\). Choose the proposal with the smallest key and accept it if \(\Psi(S')<_{\mathrm{lex}}\Psi(S)\). Stop when no sampled proposal improves the key or after \(80\) accepted exchanges. The algorithm returns the best selection encountered under the feasibility-first comparison rule. The ranking constraint was satisfied in all 45 runs of the 20\% retention experiment across 15 benchmarks and three seeds.

\paragraph{Properties of refinement.}
\label{app:refinement-properties}

The reference-bank quantile guarantees at least one feasible reference subset, so feasibility-first initialization starts from a feasible subset. Each accepted exchange preserves $|S|=B$ and feasibility while strictly decreasing $R_{\mathrm{mean}}(S)$. Strict decrease over the finite collection of size-$B$ subsets prevents revisits and guarantees finite termination.

\subsection{Source-Assisted Estimation}
\label{app:source-correction}
\label{app:theory-correspondence}

Algorithm~\ref{alg:constrained-selection} evaluates selection risks using uncorrected source means and applies source-assisted correction after the subset is fixed.

\paragraph{Residual Estimation Identity}
\label{app:correction-error}

After fixing \(S\), we combine the full-benchmark mean of auxiliary predictions with the mean observed target residual on \(S\). This uses the correction principle of prediction-powered mean estimation \citep{angelopoulos2023prediction}, with source-average scores serving as the auxiliary predictions. Let
\[
z_{nk}=\frac1M\sum_{m=1}^{M}X_{mnk},
\qquad \xi_{gnk}=Y_{gnk}-z_{nk}.
\]
Bars denote full-item means, and a subscript \(S\) denotes subset means. Then
\[
\begin{aligned}
\mu_{gk}&=\overline z_k+\overline \xi_{gk},\\
\widetilde\mu_{gk}(S)
&=\widehat\mu_{gk}(S)+\overline z_k-\overline z_{S,k}
=\overline z_k+\overline \xi_{gS,k}.
\end{aligned}
\]
Subtracting the full target mean gives the exact identity
\begin{equation}
\boxed{
\widetilde\mu_{gk}(S)-\mu_{gk}
=\overline \xi_{gS,k}-\overline \xi_{gk}
=\delta_{gk}(S)-\frac1M\sum_{m=1}^{M}\delta_{mk}(S)
}.
\label{eq:corrected-error-identity}
\end{equation}
The adjustment replaces the auxiliary component's subset mean with its known full-benchmark mean. The remaining error is the subset error of the target residual.

\paragraph{Exact Preservation of Model Ordering}
\label{app:correction-ranking}

\textbf{Proposition 1.} Source-assisted difference correction preserves every target-model pairwise score difference within each metric.

\textbf{Proof.} For metric $k$, the correction \(c_k=\overline z_k-\overline z_{S,k}\) is shared by all target models. Hence, for any \(a,b\),
\[
\widetilde\mu_{ak}-\widetilde\mu_{bk}
=
(\widehat\mu_{ak}+c_k)-(\widehat\mu_{bk}+c_k)
=
\widehat\mu_{ak}-\widehat\mu_{bk}.
\]

The correction preserves all within-metric orders and ties. Ranking fidelity is determined by item selection.

\subsection{Source-to-Target Error Analysis}
\label{app:error-analysis}

\subsubsection{From Source Errors to Target Errors}
\label{app:score-transfer}

Source-guided selection is useful when source and target models share item-level performance patterns. The following decomposition makes this relationship explicit.

For metric \(k\), write the target score vector as
\[
\mathbf y_{gk}
=
a_{gk}\mathbf1
+
\sum_{m=1}^{M}\alpha_{gmk}\mathbf x_{mk}
+
\mathbf e_{gk},
\]
where \(\mathbf x_{mk},\mathbf y_{gk}\in\mathbb R^N\), and \(\mathbf e_{gk}\) captures the remaining item-level variation. We use this representation to analyze transfer.

Define \(\mathbf d_S=\mathbf1_S/B-\mathbf1/N\). Then \(\mathbf d_S^\top\mathbf1=0\), and the target error satisfies
\[
\delta_{gk}(S)
=
\sum_{m=1}^{M}\alpha_{gmk}\delta_{mk}(S)
+
\mathbf d_S^\top\mathbf e_{gk}.
\]
Consequently,
\begin{equation}
\boxed{
|\delta_{gk}(S)|
\le
\sum_{m=1}^{M}|\alpha_{gmk}|\,|\delta_{mk}(S)|
+
\sqrt{\frac1B-\frac1N}\,\|\mathbf e_{gk}\|_2
}.
\label{eq:source-target-error-bound}
\end{equation}

\textbf{Proof.} Multiply the score decomposition by \(\mathbf d_S^\top\). The constant term vanishes because \(\mathbf d_S^\top\mathbf1=0\). Apply the triangle inequality and Cauchy–Schwarz inequality, together with
\[
\|\mathbf d_S\|_2^2
=
B\left(\frac1B-\frac1N\right)^2
+
(N-B)\frac1{N^2}
=
\frac1B-\frac1N.
\]

The bound separates target error into a source-error contribution and a residual contribution. When source scores capture the target's item-level variation with moderate coefficients and a small residual, controlling errors across source models directly controls the first contribution. The relation applies separately to each metric, motivating simultaneous control of source reconstruction errors across the full evaluation matrix.

\subsubsection{Score Errors and Ranking Preservation}
\label{app:ranking-need}

Ranking preservation depends on estimation errors relative to the original score gaps. Consider two models \(a\) and \(b\) under metric \(k\), with full-benchmark gap \(d_{abk}=\mu_{ak}-\mu_{bk}>0\). Their subset gap is
\[
\widehat d_{abk}(S)
=
d_{abk}+\delta_{ak}(S)-\delta_{bk}(S).
\]
Hence,
\[
|\delta_{ak}(S)|+|\delta_{bk}(S)|<d_{abk}
\quad\Longrightarrow\quad
\widehat d_{abk}(S)>0.
\]

\textbf{Derivation.}
\[
\widehat d_{abk}(S)
\ge
d_{abk}
-
|\delta_{ak}(S)|
-
|\delta_{bk}(S)|
>0.
\]

The sufficient condition depends on the errors of a particular model pair relative to its score gap. A small error averaged over all models and metrics does not ensure this pair-specific condition, especially for closely spaced models. This distinguishes overall score reconstruction from preservation of model comparisons and motivates evaluating signed pairwise margins directly. The ranking risk defined in Appendix~\ref{app:rank-risk} penalizes margin shrinkage as well as reversed orderings.

\clearpage

\section{Experimental Setup}
\label{app:experimental-setup}

Tab.~\ref{tab:benchmark_overview} summarizes the tasks and model, item, and component counts for the 15 benchmark configurations, which together provide 85 component metrics. Component scores retain their benchmark-specific definitions, and evaluation NMAE uses the held-out target-model scale in Eq.~\ref{eq:eval-nmae}.

\subsection{Baseline Implementations}
\label{app:external-baselines}

\noindent\textbf{Common protocol.}
All external baselines use the same six source models as HoloCosm.
Preprocessing statistics, hyperparameter choices, item selection, and
estimator fitting are determined using only source-model information and
method-specific item features. Target-model responses are never used
during selection or tuning; target models are evaluated only after the
selected subset and estimation procedure have been fixed.

\noindent\textbf{tinyBenchmarks.}
For continuous components, we min--max normalize source scores and
choose a binarization threshold to match the binary response mean to the
original source-score mean. We fit a five-dimensional 2PL model with
hierarchical priors and represent each item by its fitted discrimination
and difficulty parameters. Weighted $K$-means forms $B$ clusters, and
linear-sum assignment selects one distinct item per center. Cluster
frequencies provide aggregation weights.

\noindent\textbf{ZipBench.}
ZipBench uses the same preprocessing and IRT configuration as tinyBenchmarks.
We augment the six observed source profiles with six synthetic profiles
constructed by majority voting over source-model combinations across
performance levels. We cluster the fitted item representations into $B$
groups, select each center's nearest item, and weight selected items by
cluster frequency.

\noindent\textbf{mRMR.}
mRMR greedily selects items using the mutual-information quotient, with
the full-benchmark source-model mean as the relevance target. We use
appropriate mutual-information estimators for binary and continuous
components. A separate RidgeCV estimator is fitted for each evaluation
component to reconstruct target scores from the selected items.

\noindent\textbf{AnchorPoints.}
AnchorPoints computes Pearson correlations between item response patterns
across source models and applies fasterpam to the distance
$1-\mathrm{correlation}$. The medoids are selected, and each
item is assigned to its most-correlated medoid. Cluster sizes provide
aggregation weights.

\noindent\textbf{EssenceBench.}
We use BGE-M3 embeddings and size-constrained filtering, with GA/EBM search when the filtered pool exceeds the required subset size.
Selection uses leave-one-source-model-out validation and GAM-based
fitness. Target scores are estimated by averaging selected scores.

\subsection{Simple Multi-Metric Strategies}
\label{app:simple-multimetric-details}

All strategies use identical source--target splits and 20\% retention.
Evaluation uses the original metrics. We compute mean and worst-component
fidelity per run, average runs within each configuration, and weight the
15 configurations equally. For the Single-metric oracle reference, we report
the selection metric with the lowest target-set mean NMAE.

\noindent\textbf{AverageMetric.}
We align metric directions, apply source-derived min--max normalization,
and average metrics equally. Methods use it for selection and retain
existing score-estimation procedures.

\noindent\textbf{Vote.}
We combine the size-$B$ subsets selected by AnchorPoints for individual
metrics. Each selected item receives one vote per metric, and the $B$
highest-voted distinct items are retained, with balanced metric coverage
used for tie-breaking. Their unweighted means estimate all target metrics.

\noindent\textbf{K-means.}
Let $s_k$ be the standard deviation, across source models, of their
full-benchmark means on metric $k$; numerically zero scales are set to one.
For each item, we concatenate source responses normalized by $s_k$:
\[
v_n=\operatorname{vec}([X_{mnk}/s_k]_{m,k}),\qquad
\widehat{\mu}_{gk}=\sum_{j=1}^{B}\frac{|C_j|}{N}Y_{g,s_j,k}.
\]
We use Lloyd K-means with $B$ clusters and select the actual item nearest
each cluster center, enforcing distinct representatives when needed.
Cluster proportions provide the weights above, and target models evaluate
only the $B$ representatives. Detailed comparisons are reported in
Appendix~\ref{app:strategy-results}.

\clearpage

\section{Analysis and Discussion}
\label{app:analysis-discussion}

\subsection{Why Random Sampling Is a Strong Baseline}
\label{app:random-scores}

Uniform sampling estimates full-benchmark means without requiring
source-model information. Its competitiveness is also highlighted by
\citet{yauney2026reliable}, who examine the reliability of
micro-benchmarks in preserving pairwise model comparisons.

For a subset $S$ sampled uniformly without replacement from all
size-$B$ subsets, the uncorrected subset mean satisfies
\[
\mathbb E_S[\widehat\mu_{gk}(S)]=\mu_{gk},
\qquad
\operatorname{Var}_S[\widehat\mu_{gk}(S)]
=\frac{1-B/N}{B}\,v_{gk},
\]
where $v_{gk}=(N-1)^{-1}\sum_{n=1}^{N}(Y_{gnk}-\mu_{gk})^2$
is the finite-population variance of the target model's item scores.

The estimator is unbiased, so its mean squared error equals this
variance and decreases as $B$ increases. These properties help explain
why Random is a strong baseline for score reconstruction. In benchmark
compression, however, one subset is fixed and reused across model
evaluations, making the quality of that particular subset consequential.

\paragraph{Empirical variability across random seeds.}

We evaluate Bench2Drive, LongForm, and SoB-Image at retention ratios of 10\%, 20\%, 30\%, 40\%, and 50\%. For each benchmark--retention setting, we fix the source--target split and repeat item selection with 10 random seeds. Performance is first averaged across component metrics within each run, and the reported means and standard deviations are computed across the 10 seeds.

\begin{figure}[!ht]
\centering
\includegraphics[width=\linewidth]{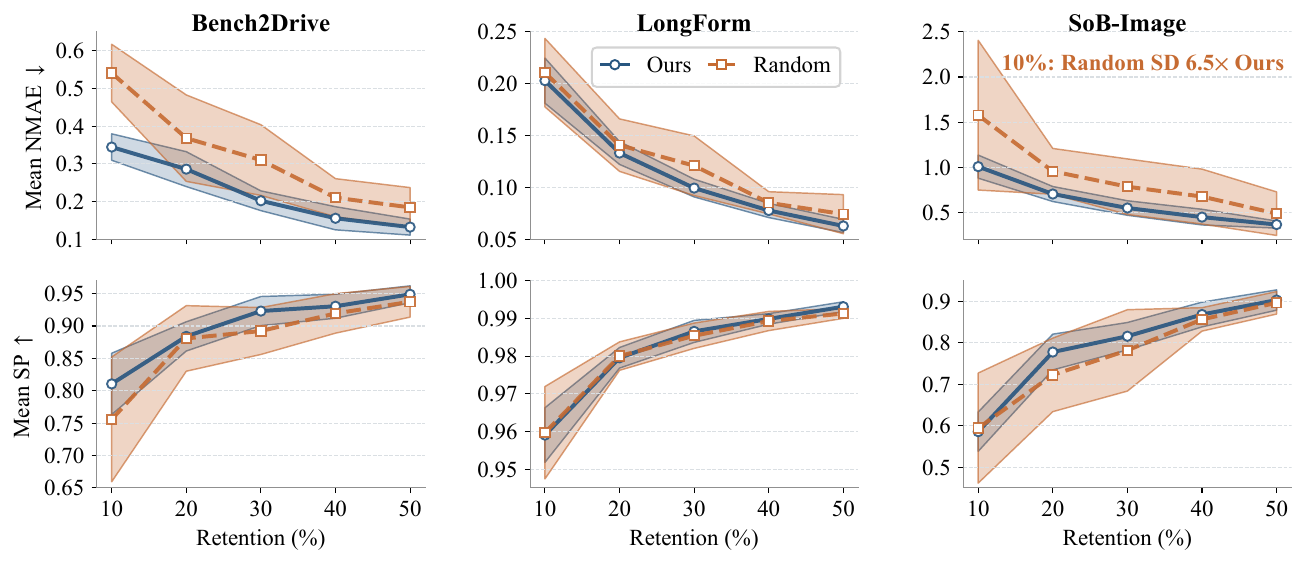}
\caption{\textbf{Random-selection variability across 10 seeds.}
Results cover Bench2Drive, LongForm, and SoB-Image at
10\%, 20\%, 30\%, 40\%, and 50\% retention under fixed model splits.
Curves show ten-seed means; shaded bands indicate one standard deviation.
Lower NMAE and higher SP indicate better fidelity.}
\label{fig:seed_stability}
\end{figure}

Fig.~\ref{fig:seed_stability} shows how evaluation fidelity
varies across random seeds at each retention level.
Thus, strong average performance can coexist with substantial variation in the quality of individual random selections. Source-guided selection addresses this practical need by using observed model responses to identify items that preserve evaluation fidelity across metrics.

\FloatBarrier

\subsection{When Is One Metric Sufficient?}
\label{app:one-metric-sufficient}

DeepResearch Bench \citep{du2026deepresearch} provides a
high-correlation counterpart to Fig.~\ref{fig:motivation}.
All ten metric pairs have model-ranking Spearman correlations of
0.92--0.99 (median 0.95), while Overall Score correlates with every
component at 0.96--0.99. We further test whether this alignment makes
single-metric selection transferable. At 20\% retention, each component
is used in turn as the sole selection metric for mRMR and AnchorPoints,
and the selected items are evaluated on all five metrics. We additionally
evaluate HoloCosm and Random at the same retention level.

\begin{figure}[!ht]
  \centering
  \includegraphics[width=0.68\linewidth]{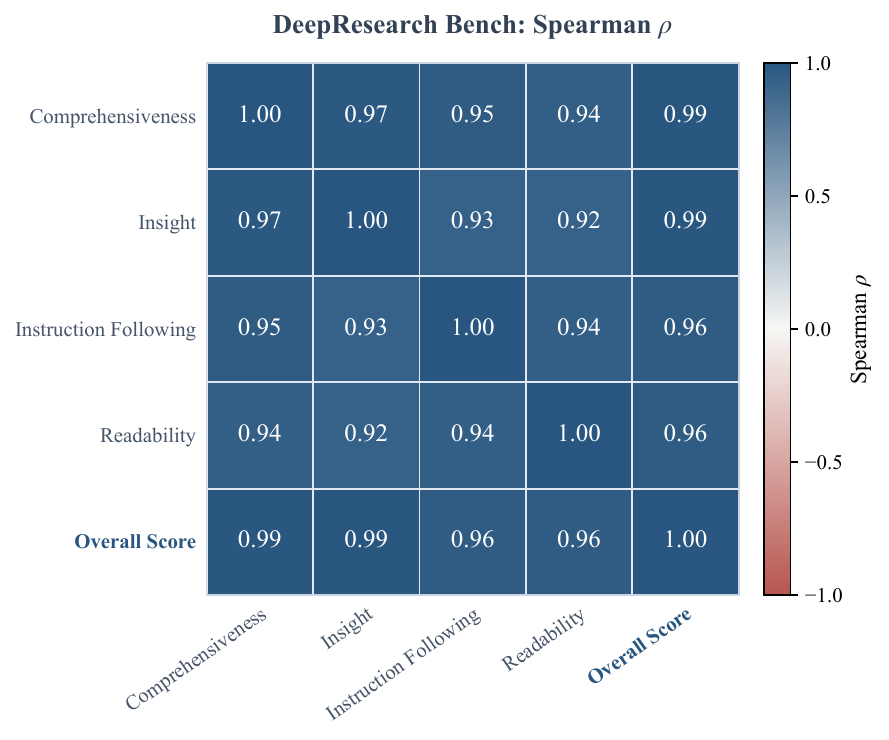}
  \caption{\textbf{A high-correlation control on DeepResearch Bench.}
  Pairwise model-ranking Spearman correlations among four component
  dimensions and Overall Score. Overall Score is highlighted.}
  \label{fig:deepresearch-correlation}
\end{figure}

\begin{table*}[t]
\setlength{\belowcaptionskip}{8pt}
\centering
\caption{\textbf{Cross-metric fidelity on DeepResearch at 20\% retention.}
Each mRMR and AnchorPoints row uses one component for selection and evaluates
the selected items on all five metrics. HoloCosm uses all metrics,
while Random uses no guiding metric. Entries report NMAE $\downarrow$ /
SP $\uparrow$, averaged over three seeds.}
\label{tab:deepresearch_cross_metric}
\scriptsize
\setlength{\tabcolsep}{3.2pt}
\resizebox{\textwidth}{!}{%
\begin{tabular}{@{}llrrrrrr@{}}
\toprule
Method & Selection metric
& Comp. & Insight & Inst.\ Follow. & Read. & Overall & Avg. \\
\midrule
HoloCosm & All metrics
& 0.12 / 0.99
& 0.10 / 0.99
& 0.21 / 0.98
& 0.10 / 0.98
& 0.10 / 0.98
& \textbf{0.13} / 0.98 \\
Random & --
& 0.10 / 0.99
& 0.09 / 0.99
& 0.27 / 0.97
& 0.15 / 0.99
& 0.09 / 0.99
& 0.14 / \textbf{0.99} \\
\midrule
\multirow{5}{*}{mRMR}
& Comprehensiveness
& 0.13 / 0.98
& 0.08 / 0.99
& 0.26 / 0.98
& 0.22 / 0.98
& 0.11 / 0.99
& 0.16 / 0.98 \\
& Insight
& 0.10 / 0.98
& 0.08 / 0.99
& 0.32 / 0.94
& 0.11 / 0.98
& 0.09 / 0.98
& 0.14 / 0.98 \\
& Instruction Following
& 0.15 / 0.99
& 0.09 / 0.99
& 0.30 / 0.94
& 0.18 / 0.98
& 0.10 / 1.00
& 0.16 / 0.98 \\
& Readability
& 0.11 / 0.98
& 0.08 / 0.99
& 0.27 / 0.96
& 0.13 / 0.96
& 0.10 / 0.99
& 0.14 / 0.98 \\
& Overall Score
& 0.11 / 0.97
& 0.09 / 0.99
& 0.22 / 0.96
& 0.17 / 0.96
& 0.11 / 0.98
& 0.14 / 0.97 \\
\midrule
\multirow{5}{*}{AnchorPoints}
& Comprehensiveness
& 0.14 / 0.98
& 0.18 / 0.99
& 0.36 / 0.96
& 0.19 / 0.98
& 0.18 / 0.99
& 0.21 / 0.98 \\
& Insight
& 0.13 / 0.99
& 0.09 / 0.99
& 0.23 / 0.96
& 0.12 / 0.99
& 0.11 / 0.99
& 0.13 / 0.98 \\
& Instruction Following
& 0.15 / 0.99
& 0.11 / 0.99
& 0.22 / 0.96
& 0.14 / 0.99
& 0.13 / 0.99
& 0.15 / 0.98 \\
& Readability
& 0.12 / 0.99
& 0.17 / 0.98
& 0.31 / 0.95
& 0.11 / 0.99
& 0.13 / 0.98
& 0.17 / 0.98 \\
& Overall Score
& 0.12 / 0.98
& 0.12 / 0.99
& 0.28 / 0.96
& 0.12 / 0.99
& 0.11 / 0.98
& 0.15 / 0.98 \\
\bottomrule
\end{tabular}%
}
\end{table*}

Regardless of the component used for selection, mRMR achieves an average
SP of 0.9714--0.9829 and AnchorPoints achieves 0.9786--0.9842 across all
evaluation metrics. Their average NMAE ranges are 0.1400--0.1645 and
0.1348--0.2087, respectively. HoloCosm's multi-metric selection reaches
0.1254/0.9817 in average NMAE/SP. These results show that when component
rankings are highly aligned, items selected using any single metric
already transfer well to the remaining metrics.

\subsection{Alternative Multi-Metric Strategies}
\label{app:alternative-strategies}

We compare AverageMetric, Vote, and K-means at 20\% retention.
Tab.~\ref{tab:simple_multimetric} summarizes the comparison;
Appendix~\ref{app:strategy-results} reports the per-benchmark results.

\subsubsection{Benchmark-Defined Overall Scores}
\label{app:official-overall}

We examine whether benchmark-level aggregate scores provide effective
guidance for preserving component metrics, using AverageMetric as the
reference. All selected subsets are evaluated on the original component
metrics at 20\% retention.

Tab.~\ref{tab:official-five} compares aggregation strategies on ConStory,
LiveIdea-Core3/Full4, and MMDocRAG Text/Multimodal.
LiveIdea-Core3 uses the average of its three included metrics, and
ConStory uses CED responses with full-data normalization.
Tab.~\ref{tab:official-eight} extends the baseline comparison to
Bench2Drive and Fail2Drive-ID/OOD, using driving\_score and HM,
respectively. Results are averaged over the configurations included
in each comparison.

Benchmark-level aggregate guidance does not consistently improve
component fidelity. HoloCosm achieves the lowest mean NMAE in both
comparisons.

\begin{table}[t]
\centering
\caption{Aggregate-guided compression at 20\% retention on five existing configurations. Official Overall includes the restricted LiveIdea-Core3 average described in the text.}
\label{tab:official-five}
\small
\begin{tabular*}{\linewidth}{@{\extracolsep{\fill}}llrrrr@{}}
\toprule
Method & Strategy & Mean NMAE & Worst NMAE & Mean SP & Worst SP \\
\midrule
AnchorPoints & AverageMetric & 0.2955 & 0.4285 & 0.9137 & 0.8613 \\
AnchorPoints & Official Overall & 0.3781 & 0.6402 & 0.9142 & 0.8707 \\
Ours & All metrics & 0.1573 & 0.2420 & 0.9561 & 0.9006 \\
Ours & AverageMetric & 0.1809 & 0.3055 & 0.9530 & 0.8942 \\
Ours & Official Overall & 0.1729 & 0.2679 & 0.9534 & 0.9009 \\
Random & Unguided & 0.1747 & 0.2813 & 0.9549 & 0.9199 \\
ZipBench & AverageMetric & 0.2802 & 0.4487 & 0.9271 & 0.8735 \\
ZipBench & Official Overall & 0.2667 & 0.4074 & 0.9152 & 0.8478 \\
mRMR & AverageMetric & 0.3859 & 0.4718 & 0.8519 & 0.7727 \\
mRMR & Official Overall & 0.3772 & 0.4801 & 0.8191 & 0.7134 \\
tinyBenchmarks & AverageMetric & 0.2656 & 0.4009 & 0.9224 & 0.8544 \\
tinyBenchmarks & Official Overall & 0.2788 & 0.4380 & 0.9220 & 0.8717 \\
\bottomrule
\end{tabular*}
\end{table}

\begin{table}[t]
\centering
\caption{Aggregate-guided compression at 20\% retention on eight configurations with common coverage. Official Overall includes the restricted LiveIdea-Core3 average described in the text.}
\label{tab:official-eight}
\small
\begin{tabular*}{\linewidth}{@{\extracolsep{\fill}}llrrrr@{}}
\toprule
Method & Strategy & Mean NMAE & Worst NMAE & Mean SP & Worst SP \\
\midrule
AnchorPoints & AverageMetric & 0.3951 & 0.5538 & 0.8756 & 0.8200 \\
AnchorPoints & Official Overall & 0.5118 & 0.7924 & 0.8454 & 0.7867 \\
Ours & All metrics & 0.2156 & 0.3050 & 0.9105 & 0.8582 \\
Ours & AverageMetric & 0.2711 & 0.3878 & 0.8845 & 0.8282 \\
Random & Unguided & 0.3295 & 0.4558 & 0.8748 & 0.8215 \\
ZipBench & AverageMetric & 0.4429 & 0.6315 & 0.8030 & 0.7382 \\
ZipBench & Official Overall & 0.4082 & 0.5670 & 0.8167 & 0.7392 \\
mRMR & AverageMetric & 0.4052 & 0.4886 & 0.7616 & 0.6692 \\
mRMR & Official Overall & 0.4519 & 0.6511 & 0.7345 & 0.5891 \\
tinyBenchmarks & AverageMetric & 0.4244 & 0.5924 & 0.8360 & 0.7539 \\
tinyBenchmarks & Official Overall & 0.4245 & 0.5850 & 0.8179 & 0.7416 \\
\bottomrule
\end{tabular*}
\end{table}

\subsection{Source-Model Selection}
\label{app:source-model-selection}

\subsubsection{Number of Source Models}
\label{app:source_count}

We vary the number of source models from 3 to 15 on LiveIdea-Full4
at 20\% retention, using the same ten target models throughout.
The remaining 31 models form the source candidate pool.
We rank these candidates by their average metric-wise
min--max-normalized full-benchmark scores and select 15 evenly
spaced representatives. For each source count $m$, we choose
$m$ evenly spaced representatives, including both endpoints,
and use the same selection across seeds.
Results are averaged over three random seeds.

\begin{figure}[H]
\centering
\includegraphics[width=\linewidth]{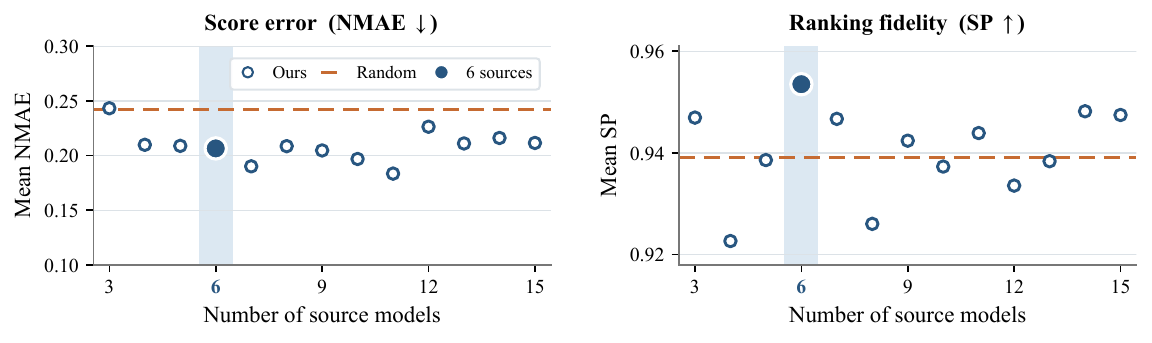}
\caption{\textbf{Source-model count sensitivity on LiveIdea-Full4.}
Results at 20\% retention are averaged over three random seeds.
All configurations use the same target models.
Lower NMAE and higher SP indicate better fidelity.}
\label{fig:source_count_liveidea}
\end{figure}

Fig.~\ref{fig:source_count_liveidea} shows that six source models
achieve the highest SP among the tested source counts.
With six sources, NMAE is 0.2065 and SP is 0.9535, compared with
0.2115 and 0.9475 with fifteen sources.
These results indicate that a compact source cohort can support
accurate multi-metric compression in this setting.

\subsubsection{Source Selection Strategy}
\label{app:source_cohort_robustness}

We compare two strategies for selecting six source models on
SoB-Audio at 20\% retention. Our strategy selects models at evenly
spaced positions in the source candidate pool's performance ranking.
\textit{Random-6} samples six models uniformly without replacement
from the same pool. Both strategies are evaluated on the same
nine target models, with results averaged over three seeds.
\begin{table}[t]
\setlength{\belowcaptionskip}{8pt}
\caption{\textbf{Random-6 versus diverse source models on SoB-Audio at 20\%
retention.} Results are averaged over three seeds.}
\label{tab:source_cohort_robustness}
\centering
\small
\setlength{\tabcolsep}{4pt}
\begin{tabular}{@{}lrrrr@{}}
\toprule
& \multicolumn{2}{c}{NMAE $\downarrow$}
& \multicolumn{2}{c}{SP $\uparrow$} \\
\cmidrule(lr){2-3}\cmidrule(lr){4-5}
Metric & Random-6 & HoloCosm & Random-6 & HoloCosm \\
\midrule
Hierarchical path recall & 0.6434 & \textbf{0.5412} & 0.7500 & \textbf{0.8389} \\
Leaf-value EM            & 0.9410 & \textbf{0.6525} & 0.5333 & \textbf{0.7944} \\
Path-set F1              & 0.7366 & \textbf{0.5841} & 0.6389 & \textbf{0.8444} \\
Schema compliance        & 0.7012 & \textbf{0.5230} & 0.7608 & \textbf{0.8507} \\
Value-token F1           & 0.5520 & \textbf{0.4463} & 0.7667 & \textbf{0.9333} \\
\midrule
Mean                     & 0.7148 & \textbf{0.5494} & 0.6899 & \textbf{0.8524} \\
\bottomrule
\end{tabular}
\end{table}

Tab.~\ref{tab:source_cohort_robustness} shows that selecting source
models across the performance ranking improves both NMAE and SP
on all five component metrics. Mean NMAE decreases from 0.7148
to 0.5494, a reduction of 23.1\%, while mean SP increases from
0.6899 to 0.8524.

\subsection{Selection and Estimation Ablations}
\label{app:additional-ablation}

We compare two additional variants at 20\% retention across all 15
configurations, using the same source--target splits, seeds, and
source-assisted estimator as in
Tab.~\ref{tab:additional_ablation}.

\paragraph{Effect of swap refinement.}
\textit{Without swap refinement} uses the same 128 reference subsets and
ranking threshold as HoloCosm, selects the feasible reference subset
with the lowest reconstruction error, and stops without exchanging items.
This isolates the contribution of refinement from reference-bank selection.
As shown in Tab.~\ref{tab:search-objective-ablation}, refinement further
reduces Mean NMAE by 5.53\%, with a smaller increase in Mean SP.

\paragraph{Reconstruction objective.}
\textit{Corrected-residual objective} replaces only the mean reconstruction
loss, retaining the source-pool folds, metric scales, ranking constraint,
and search budget. Within each fold, each source model is treated as a
pseudo-target and corrected using the other models in that fold.
For a fold $F$ of size $L$, let $\delta_{mk}(S)$ be model $m$'s
uncorrected subset error. The corrected error is
\[
e^{(-m)}_{mk}(S)
= \delta_{mk}(S)-\frac{1}{L-1}\sum_{j\in F\setminus\{m\}}\delta_{jk}(S)
= \frac{L}{L-1}\left(\delta_{mk}(S)-\overline\delta_{F,k}(S)\right).
\]
We average its normalized absolute value across models, metrics, and
the existing folds. This alternative increases mean NMAE by 19.29\% and lowers
mean SP relative to HoloCosm.

\begin{table}[t]
\centering
\caption{\textbf{Swap refinement and reconstruction-objective ablations
at 20\% retention.} All variants use matched seeds and the same
source-assisted estimator. Results are averaged equally over 15
configurations.}
\label{tab:search-objective-ablation}
\small
\begin{tabular*}{\linewidth}{@{\extracolsep{\fill}}lrrr@{}}
\toprule
Variant & Mean NMAE $\downarrow$ & Worst NMAE $\downarrow$ & Mean SP $\uparrow$ \\
\midrule
Without swap refinement & 0.2500 & 0.3544 & 0.9130 \\
Corrected-residual objective & 0.2817 & 0.3802 & 0.8983 \\
HoloCosm & \textbf{0.2362} & \textbf{0.3299} & \textbf{0.9151} \\
\bottomrule
\end{tabular*}
\end{table}

Tab.~\ref{tab:fail2drive_ablation} isolates multi-metric selection and
pairwise-ranking protection on Fail2Drive-ID. \textit{Single-metric
selection} uses HM, DR, and SR in turn to guide selection and averages the
results over guiding metrics. \textit{Without ranking constraint} removes
only the pairwise-ranking constraint. Both ablations degrade all three
metrics: multi-metric selection reduces average NMAE by 35.1\% over single-metric
selection, while removing the constraint changes NMAE/SP from
0.2762/0.7964 to 0.3573/0.6594.

\begin{table}[!ht]
\setlength{\belowcaptionskip}{8pt}
\centering
\caption{\textbf{Ablation on Fail2Drive-ID at 20\% retention.}
Three-seed NMAE $\downarrow$ / SP $\uparrow$ means; Avg.\ averages evaluation
metrics. Single-metric results also average guiding metrics.}
\label{tab:fail2drive_ablation}
\small
\setlength{\tabcolsep}{4pt}
\begin{tabular}{lcccc}
\toprule
Variant & HM & DR & SR & Avg. \\
\midrule
Ours
& \textbf{0.2940} / \textbf{0.7845}
& \textbf{0.2246} / \textbf{0.8345}
& \textbf{0.3101} / \textbf{0.7703}
& \textbf{0.2762} / \textbf{0.7964} \\
Single-metric selection
& 0.4507 / 0.6663
& 0.3309 / 0.6968
& 0.4960 / 0.6636
& 0.4259 / 0.6756 \\
Without ranking constraint
& 0.3710 / 0.6429
& 0.3099 / 0.6714
& 0.3911 / 0.6640
& 0.3573 / 0.6594 \\
\bottomrule
\end{tabular}
\end{table}

Tabs.~\ref{tab:workbench_constraint_per_metric} and
\ref{tab:liveidea_constraint_per_metric} resolve this comparison at the
component level on two three-metric benchmarks. On WorkBench, Ours reduces
NMAE on all three metrics, with the largest reduction on Exact Match
(16.5\%). On LiveIdea-Core3, the reduction is concentrated on Feasibility
(16.7\%), while Clarity and Originality change by less than 3\%.

\begin{table}[!ht]
\setlength{\belowcaptionskip}{8pt}
\centering
\caption{\textbf{Per-metric constrained-selection ablation on WorkBench.}
Results at 20\% retention are three-run means. Reconstruction and Ours use the same
source-assisted correction. NMAE reduction is $(\mathrm{Recon}-\mathrm{Ours})/
\mathrm{Recon}$; SP gain is $\mathrm{Ours}-\mathrm{Recon}$. Positive values
favor Ours.}
\label{tab:workbench_constraint_per_metric}
\small
\setlength{\tabcolsep}{2.8pt}
\begin{tabular}{@{}lrrrrrr@{}}
\toprule
& \multicolumn{3}{c}{NMAE $\downarrow$} & \multicolumn{3}{c}{SP $\uparrow$} \\
\cmidrule(lr){2-4}\cmidrule(lr){5-7}
Metric & Recon & Ours & Reduction & Recon & Ours & Gain \\
\midrule
Correct     & 0.1384 & \textbf{0.1288} & 6.9\%  & 0.9790 & \textbf{0.9855} & +0.0065 \\
Exact Match & 0.1561 & \textbf{0.1304} & 16.5\% & 0.9814 & \textbf{0.9837} & +0.0022 \\
Safety      & 0.1696 & \textbf{0.1578} & 6.9\%  & 0.9592 & \textbf{0.9614} & +0.0022 \\
\bottomrule
\end{tabular}
\end{table}

\begin{table}[!ht]
\setlength{\belowcaptionskip}{8pt}
\centering
\caption{\textbf{Per-metric constrained-selection ablation on LiveIdea-Core3.}
Results at 20\% retention are three-run means. Reconstruction and Ours use the same
source-assisted correction. NMAE reduction is $(\mathrm{Recon}-\mathrm{Ours})/
\mathrm{Recon}$; SP gain is $\mathrm{Ours}-\mathrm{Recon}$. Positive values
favor Ours.}
\label{tab:liveidea_constraint_per_metric}
\small
\setlength{\tabcolsep}{2.8pt}
\begin{tabular}{@{}lrrrrrr@{}}
\toprule
& \multicolumn{3}{c}{NMAE $\downarrow$} & \multicolumn{3}{c}{SP $\uparrow$} \\
\cmidrule(lr){2-4}\cmidrule(lr){5-7}
Metric & Recon & Ours & Reduction & Recon & Ours & Gain \\
\midrule
Clarity     & \textbf{0.1054} & 0.1074 & $-1.9$\% & 0.9725 & \textbf{0.9738} & +0.0012 \\
Feasibility & 0.1564 & \textbf{0.1303} & 16.7\%   & 0.9641 & \textbf{0.9771} & +0.0130 \\
Originality & \textbf{0.0690} & 0.0707 & $-2.6$\% & \textbf{0.9947} & \textbf{0.9947} & 0.0000 \\
\bottomrule
\end{tabular}
\end{table}

\subsubsection{Comparison under a Common Estimator}
\label{app:baseline-correction}

We compare HoloCosm with AverageMetric-guided ZipBench under the same
uniform source-assisted estimator at 20\% retention across all 15
configurations. We also report ZipBench's original weighted estimator.
The ZipBench variants use the same selected items, and all estimates
are evaluated on the original component metrics.

\begin{table}[!htbp]
\setlength{\belowcaptionskip}{8pt}
\centering
\caption{\textbf{Common-estimator comparison with AverageMetric-guided
ZipBench at 20\% retention.} Results average three runs per method
and weight the 15 benchmarks equally. All ZipBench variants use
the same selected items.}
\label{tab:common-estimator}
\small
\setlength{\tabcolsep}{4pt}
\begin{tabular}{@{}lrrr@{}}
\toprule
Estimator / method & Mean NMAE $\downarrow$ & Worst NMAE $\downarrow$ & Mean SP $\uparrow$ \\
\midrule
ZipBench: original weighted & 0.4920 & 0.7376 & 0.8145 \\
ZipBench: uniform + correction & 0.3423 & 0.4720 & 0.8705 \\
HoloCosm & \textbf{0.2362} & \textbf{0.3299} & \textbf{0.9151} \\
\bottomrule
\end{tabular}
\end{table}

Tab.~\ref{tab:common-estimator} shows that, under the same uniform
source-corrected estimator, HoloCosm-selected subsets achieve
31.0\% lower mean NMAE and 30.1\% lower worst-component NMAE than
AverageMetric-guided ZipBench subsets. Mean NMAE is lower on
13 of the 15 configurations.

\subsubsection{Sensitivity to the Ranking-Constraint Quantile}
\label{app:hyperparameter-sensitivity}

We evaluate ranking-constraint calibration quantiles
$q\in\{0.10,0.25,0.50\}$ on Fail2Drive-ID, SoB-Audio, and
LiveIdea-Core3 at 20\% retention.
Tab.~\ref{tab:hyperparameter-sensitivity} reports results
averaged over three seeds.

\begin{table}[!ht]
\setlength{\belowcaptionskip}{8pt}
\centering
\caption{\textbf{Sensitivity to the ranking-constraint quantile at 20\% retention.}
Each entry reports mean/worst-component NMAE / mean SP.}
\label{tab:hyperparameter-sensitivity}
\small
\setlength{\tabcolsep}{2.5pt}
\begin{tabular}{@{}lccc@{}}
\toprule
Setting & Fail2Drive-ID & SoB-Audio & LiveIdea-Core3 \\
\midrule
$q=0.10$
& 0.3996 / 0.4671 / 0.5782
& 0.6236 / 0.7902 / 0.7960
& 0.1028 / 0.1336 / 0.9819 \\
$q=0.25$ (default)
& \textbf{0.2762 / 0.3219 / 0.7964}
& \textbf{0.5798 / 0.7520 / 0.8293}
& \textbf{0.1028 / 0.1336 / 0.9819} \\
$q=0.50$
& 0.3573 / 0.3922 / 0.6594
& 0.5798 / 0.7520 / 0.8293
& 0.1102 / 0.1564 / 0.9771 \\
\bottomrule
\end{tabular}
\end{table}

The default $q=0.25$ achieves the best overall fidelity
across the three benchmarks.

\subsection{Evaluation Cost}
\label{app:evaluation-cost}

\subsubsection{Evaluation Size}
\label{app:metric-wise-evaluation-cost}
\label{app:union-results}

For metric-wise selection, the nominal per-metric selection size need not equal the
final evaluation cost. Let $S_k(B)$ denote the selection for component $k$. The union $S_{\mathrm{union}}(B)=\bigcup_{k=1}^{K}S_k(B)$ contains between $B$ and $\min(N,KB)$ unique items. For each benchmark,
method, retention level, and seed, the evaluation cost is the number of distinct
items in this union, with items shared by several component selections counted
only once.

We report the inflation factor $I_{\mathrm{union}}=|S_{\mathrm{union}}(B)|/B$.
Tab.~\ref{tab:union_size_examples} gives representative tinyBenchmarks selection
sizes and absolute union sizes. For example, on LongForm a per-component
selection size of 22 items produces a union of 101 items, or 4.591 times the
nominal selection size.

\begin{table}[!htbp]
\setlength{\belowcaptionskip}{8pt}
\caption{\textbf{Representative actual union sizes for tinyBenchmarks.}
At 20\% nominal retention, independent component-wise selections can
require substantially more than $B$ unique evaluations.}
\label{tab:union_size_examples}
\centering
\small
\setlength{\tabcolsep}{6pt}
\begin{tabular}{@{}lrrr@{}}
\toprule
Benchmark & $B$ & $|S_{\mathrm{union}}|$ & $I_{\mathrm{union}}$ \\
\midrule
LongForm & 22 & 101 & 4.591 \\
Widget2Code & 58 & 263 & 4.534 \\
MMDocRAG-MM & 395 & 1,624 & 4.111 \\
SoB-Text & 1,000 & 3,320 & 3.320 \\
RouterArena & 1,680 & 3,020 & 1.798 \\
\bottomrule
\end{tabular}

\end{table}

\FloatBarrier

\subsubsection{Offline Item-Selection Time}
\label{app:construction-time}

At 20\% retention, item selection averages 0.77 seconds across all
45 benchmark--seed runs. Tab.~\ref{tab:construction_time} gives results for
three larger benchmarks; the longest individual run is 3.71 seconds on
RouterArena.

\begin{table}[H]
\setlength{\belowcaptionskip}{8pt}
\caption{\textbf{Offline item-selection time at 20\% retention.}
Times are means and one standard deviation over three seeds.}
\label{tab:construction_time}
\centering
\small
\setlength{\tabcolsep}{8pt}
\begin{tabular}{@{}lrrrr@{}}
\toprule
Benchmark & Items & Metrics & Selection size & Time (s) \\
\midrule
RouterArena & 8,400 & 2 & 1,680 & $2.53 \pm 1.12$ \\
SoB-Text & 5,000 & 6 & 1,000 & $2.38 \pm 0.28$ \\
MMDocRAG-Multimodal & 1,974 & 8 & 395 & $1.32 \pm 0.17$ \\
\bottomrule
\end{tabular}
\end{table}

\FloatBarrier
\Needspace{28\baselineskip}
\raggedbottom

\section{Detailed Results}
\label{app:results}

\subsection{Aggregate Results}
\label{app:all-target-results}

Tab.~\ref{tab:all_target_summary} reports the aggregate fidelity measures.
We additionally report Kendall's $\tau_b$, which accounts for tied
ranks~\citep{kendall1945treatment}.
For each single-target method and benchmark, every official component is
used as a selection target, and the selected items are evaluated on all
components. Component means and extrema are computed for each selection before
averaging over repetitions. Each baseline retains the target with the lowest
mean NMAE and reports all fidelity measures for that target. The final summary weights
the 15 benchmarks equally. HoloCosm and Random each perform one selection per
run rather than metric-specific selection.

\begin{table}[H]
\setlength{\belowcaptionskip}{8pt}
\caption{\textbf{Aggregate fidelity across 15 benchmarks at 20\% retention.}
Mean and worst-component fidelity are aggregated within benchmarks before
averaging benchmarks equally. Each scalar method uses its best selection
target by mean NMAE, with the same target used for all fidelity measures. Bold and underline indicate the best and
second-best values in each column.}
\label{tab:all_target_summary}
\centering
\small
\setlength{\tabcolsep}{5pt}
\begin{tabular}{@{}lrrrrr@{}}
\toprule
& \multicolumn{2}{c}{NMAE $\downarrow$} & \multicolumn{2}{c}{SP $\uparrow$} & Kendall\textquotesingle s $\tau_b\uparrow$ \\
\cmidrule(lr){2-3}\cmidrule(lr){4-5}
Method & Mean & Worst & Mean & Worst & Mean \\
\midrule
HoloCosm & \textbf{0.2362} & \textbf{0.3299} & \textbf{0.9151} & \textbf{0.8624} & \textbf{0.8173} \\
Random & \underline{0.3650} & \underline{0.5450} & \underline{0.8840} & \underline{0.8302} & \underline{0.7859} \\
tinyBenchmarks & 0.4421 & 0.6297 & 0.8483 & 0.7545 & 0.7252 \\
ZipBench & 0.4123 & 0.6364 & 0.8467 & 0.7646 & 0.7288 \\
mRMR & 0.4321 & 0.5997 & 0.7894 & 0.5952 & 0.6734 \\
AnchorPoints & 0.5533 & 0.9256 & 0.8361 & 0.7508 & 0.7196 \\
EssenceBench & 0.4515 & 0.6302 & 0.8736 & 0.7898 & 0.7643 \\
\bottomrule
\end{tabular}

\end{table}

\Needspace{22\baselineskip}
\begingroup
\setlength{\intextsep}{4pt}
\subsection{Per-Benchmark Strategy Comparisons}
\label{app:strategy-results}

Tabs.~\ref{tab:simple-details-a} and~\ref{tab:simple-details-b} expand
Tab.~\ref{tab:simple_multimetric}. All values are averages over available
runs within each configuration; Average weights configurations equally.

\begin{table}[H]
\centering
\caption{Per-configuration simple-strategy results at 20\% retention: Random and AnchorPoints. NMAE and SP are means over metrics.}
\label{tab:simple-details-a}
\small
\resizebox{\linewidth}{!}{%
\begin{tabular}{lrrrrrrrr}
\toprule
Configuration & \multicolumn{2}{c}{Random} & \multicolumn{2}{c}{AnchorPoints: Single} & \multicolumn{2}{c}{AnchorPoints: AverageMetric} & \multicolumn{2}{c}{AnchorPoints: Vote} \\
 & NMAE $\downarrow$ & SP $\uparrow$ & NMAE $\downarrow$ & SP $\uparrow$ & NMAE $\downarrow$ & SP $\uparrow$ & NMAE $\downarrow$ & SP $\uparrow$ \\
\midrule
Widget2Code & 0.2979 & 0.9335 & 0.3280 & 0.8628 & 0.3148 & 0.9167 & 0.4707 & 0.8911 \\
ConStory & 0.2938 & 0.8813 & 1.2431 & 0.5209 & 0.8375 & 0.7038 & 0.3320 & 0.8606 \\
Bench2Drive & 0.3881 & 0.8418 & 0.5910 & 0.7899 & 0.5200 & 0.8120 & 0.9476 & 0.8800 \\
Fail2Drive-ID & 0.5805 & 0.5918 & 0.2839 & 0.7051 & 0.4309 & 0.7524 & 0.8388 & 0.8696 \\
Fail2Drive-OOD & 0.7939 & 0.7903 & 0.6111 & 0.7968 & 0.7323 & 0.8720 & 0.7760 & 0.7794 \\
LiveIdea-Core3 & 0.1075 & 0.9834 & 0.1207 & 0.9811 & 0.1182 & 0.9733 & 0.1107 & 0.9827 \\
LiveIdea-Full4 & 0.2888 & 0.9359 & 0.2693 & 0.9410 & 0.3016 & 0.9225 & 0.2539 & 0.9490 \\
LongForm & 0.1400 & 0.9814 & 0.1459 & 0.9801 & 0.1483 & 0.9780 & 0.1546 & 0.9738 \\
MMDocRAG-Multimodal & 0.0729 & 0.9915 & 0.0864 & 0.9882 & 0.0886 & 0.9908 & 0.1494 & 0.9883 \\
MMDocRAG-Text & 0.1106 & 0.9824 & 0.1259 & 0.9740 & 0.1318 & 0.9780 & 0.3201 & 0.9583 \\
RouterArena & 0.1464 & 0.9905 & 0.2783 & 0.9714 & 0.5413 & 0.9048 & 0.8428 & 0.9857 \\
SoB-Audio & 0.7048 & 0.7791 & 0.7179 & 0.8040 & 1.3456 & 0.6978 & 1.3263 & 0.7372 \\
SoB-Image & 1.1731 & 0.6484 & 0.8664 & 0.7090 & 1.0741 & 0.7303 & 1.8610 & 0.7579 \\
SoB-Text & 0.2104 & 0.9598 & 1.8020 & 0.6720 & 1.8754 & 0.7646 & 0.8672 & 0.9657 \\
WorkBench & 0.1664 & 0.9683 & 0.8299 & 0.8448 & 0.6982 & 0.7484 & 0.1812 & 0.9775 \\
\midrule
Average & 0.3650 & 0.8840 & 0.5533 & 0.8361 & 0.6106 & 0.8497 & 0.6288 & 0.9038 \\
\bottomrule
\end{tabular}}
\end{table}

\begin{table}[H]
\centering
\caption{Per-configuration simple-strategy results at 20\% retention: K-means and Ours. NMAE and SP are means over metrics.}
\label{tab:simple-details-b}
\small
\resizebox{0.80\linewidth}{!}{%
\begin{tabular}{lrrrrrr}
\toprule
Configuration & \multicolumn{2}{c}{K-means} & \multicolumn{2}{c}{Ours: AverageMetric} & \multicolumn{2}{c}{Ours: All metrics} \\
 & NMAE $\downarrow$ & SP $\uparrow$ & NMAE $\downarrow$ & SP $\uparrow$ & NMAE $\downarrow$ & SP $\uparrow$ \\
\midrule
Widget2Code & 0.2231 & 0.9196 & 0.2086 & 0.9276 & 0.2178 & 0.9444 \\
ConStory & 0.5427 & 0.7896 & 0.4015 & 0.8618 & 0.2972 & 0.8754 \\
Bench2Drive & 0.2938 & 0.8346 & 0.3234 & 0.8677 & 0.2736 & 0.8432 \\
Fail2Drive-ID & 0.2370 & 0.7865 & 0.3761 & 0.7031 & 0.2762 & 0.7964 \\
Fail2Drive-OOD & 0.4327 & 0.7447 & 0.5646 & 0.7401 & 0.3889 & 0.8642 \\
LiveIdea-Core3 & 0.1425 & 0.9738 & 0.1056 & 0.9811 & 0.1028 & 0.9819 \\
LiveIdea-Full4 & 0.3216 & 0.9335 & 0.2414 & 0.9471 & 0.2247 & 0.9481 \\
LongForm & 0.1474 & 0.9744 & 0.1355 & 0.9763 & 0.1172 & 0.9839 \\
MMDocRAG-Multimodal & 0.0756 & 0.9903 & 0.0658 & 0.9917 & 0.0645 & 0.9929 \\
MMDocRAG-Text & 0.1411 & 0.9702 & 0.0902 & 0.9832 & 0.0972 & 0.9821 \\
RouterArena & 0.2161 & 0.9524 & 0.1207 & 0.9810 & 0.0928 & 0.9714 \\
SoB-Audio & 0.6312 & 0.8668 & 0.5812 & 0.7927 & 0.5798 & 0.8293 \\
SoB-Image & 0.7921 & 0.6235 & 0.7079 & 0.7217 & 0.5145 & 0.7723 \\
SoB-Text & 0.3489 & 0.9415 & 0.1649 & 0.9692 & 0.1560 & 0.9642 \\
WorkBench & 0.3171 & 0.9620 & 0.1463 & 0.9753 & 0.1390 & 0.9769 \\
\midrule
Average & 0.3242 & 0.8842 & 0.2822 & 0.8946 & 0.2362 & 0.9151 \\
\bottomrule
\end{tabular}}
\end{table}

\Needspace{16\baselineskip}
Tab.~\ref{tab:overall_all_benchmarks_detailed} reports additional baseline results using AverageMetric for item selection at 20\% retention.

\begin{table}[H]
\centering
\caption{Additional AverageMetric-guided results at 20\% retention. Ours: All metrics uses the full metric set. NMAE and SP are metric means; Average weights 15 configurations equally.}
\label{tab:overall_all_benchmarks_detailed}
\small
\resizebox{\linewidth}{!}{%
\begin{tabular}{lrrrrrrrrrrrr}
\toprule
Configuration & \multicolumn{2}{c}{tinyBenchmarks} & \multicolumn{2}{c}{ZipBench} & \multicolumn{2}{c}{mRMR} & \multicolumn{2}{c}{AnchorPoints} & \multicolumn{2}{c}{Ours: AverageMetric} & \multicolumn{2}{c}{Ours: All metrics} \\
 & NMAE $\downarrow$ & SP $\uparrow$ & NMAE $\downarrow$ & SP $\uparrow$ & NMAE $\downarrow$ & SP $\uparrow$ & NMAE $\downarrow$ & SP $\uparrow$ & NMAE $\downarrow$ & SP $\uparrow$ & NMAE $\downarrow$ & SP $\uparrow$ \\
\midrule
Widget2Code & 0.4120 & 0.8958 & 0.3892 & 0.8562 & 0.2476 & 0.9167 & 0.3148 & 0.9167 & 0.2086 & 0.9276 & 0.2178 & 0.9444 \\
ConStory & 0.4377 & 0.8259 & 0.5151 & 0.8100 & 0.7738 & 0.4386 & 0.8375 & 0.7038 & 0.4015 & 0.8618 & 0.2972 & 0.8754 \\
Bench2Drive & 0.7194 & 0.7078 & 0.6091 & 0.7980 & 0.5039 & 0.7149 & 0.5200 & 0.8120 & 0.3234 & 0.8677 & 0.2736 & 0.8432 \\
Fail2Drive-ID & 0.6051 & 0.7966 & 0.6112 & 0.5426 & 0.3388 & 0.5753 & 0.4309 & 0.7524 & 0.3761 & 0.7031 & 0.2762 & 0.7964 \\
Fail2Drive-OOD & 0.7430 & 0.5717 & 0.9218 & 0.4478 & 0.4691 & 0.5431 & 0.7323 & 0.8720 & 0.5646 & 0.7401 & 0.3889 & 0.8642 \\
LiveIdea-Core3 & 0.2125 & 0.9534 & 0.2012 & 0.9672 & 0.3742 & 0.9657 & 0.1182 & 0.9733 & 0.1056 & 0.9811 & 0.1028 & 0.9819 \\
LiveIdea-Full4 & 0.3656 & 0.8915 & 0.4070 & 0.9139 & 0.4173 & 0.8896 & 0.3016 & 0.9225 & 0.2414 & 0.9471 & 0.2247 & 0.9481 \\
LongForm & 0.1878 & 0.9641 & 0.2170 & 0.9623 & 0.1727 & 0.9735 & 0.1483 & 0.9780 & 0.1355 & 0.9763 & 0.1172 & 0.9839 \\
MMDocRAG-Multimodal & 0.1340 & 0.9798 & 0.1110 & 0.9813 & 0.1147 & 0.9917 & 0.0886 & 0.9908 & 0.0658 & 0.9917 & 0.0645 & 0.9929 \\
MMDocRAG-Text & 0.1781 & 0.9612 & 0.1665 & 0.9631 & 0.2495 & 0.9736 & 0.1318 & 0.9780 & 0.0902 & 0.9832 & 0.0972 & 0.9821 \\
RouterArena & 0.3973 & 0.9714 & 0.3411 & 0.9048 & 1.0962 & 0.8857 & 0.5413 & 0.9048 & 0.1207 & 0.9810 & 0.0928 & 0.9714 \\
SoB-Audio & 1.0299 & 0.6290 & 1.1876 & 0.6782 & 0.5518 & 0.7420 & 1.3456 & 0.6978 & 0.5812 & 0.7927 & 0.5798 & 0.8293 \\
SoB-Image & 1.1644 & 0.6768 & 1.0201 & 0.5886 & 0.7144 & 0.2875 & 1.0741 & 0.7303 & 0.7079 & 0.7217 & 0.5145 & 0.7723 \\
SoB-Text & 0.4390 & 0.8792 & 0.4560 & 0.8417 & 0.5707 & 0.8218 & 1.8754 & 0.7646 & 0.1649 & 0.9692 & 0.1560 & 0.9642 \\
WorkBench & 0.2717 & 0.9527 & 0.2256 & 0.9612 & 0.3894 & 0.8390 & 0.6982 & 0.7484 & 0.1463 & 0.9753 & 0.1390 & 0.9769 \\
\midrule
Average & 0.4865 & 0.8438 & 0.4920 & 0.8145 & 0.4656 & 0.7706 & 0.6106 & 0.8497 & 0.2822 & 0.8946 & 0.2362 & 0.9151 \\
\bottomrule
\end{tabular}}
\end{table}

\endgroup
\FloatBarrier
\Needspace{48\baselineskip}
\subsection{Cross-Metric Results}
\label{app:small_metric_cross_results}

\setlength{\LTpre}{6pt}
\setlength{\LTpost}{6pt}
\setlength{\aboverulesep}{0.3ex}
\setlength{\belowrulesep}{0.4ex}

Tabs.~\ref{tab:cross-group-3}, \ref{tab:cross-group-3-language}, and~\ref{tab:cross-group-4} report
cross-metric fidelity at 20\% retention. Baselines select using the
indicated metric; underlining marks diagonal entries.

\begin{table}[!htbp]
\setlength{\belowcaptionskip}{8pt}
\centering
\caption{\textbf{Cross-metric fidelity on Fail2Drive-ID and Fail2Drive-OOD at 20\% retention.} Ours and Random do not use metric-specific selection. The Mean column averages NMAE across metrics; pale blue-gray marks each single-metric baseline\textquotesingle s lowest-mean selection target.}
\label{tab:cross-group-3}
\small
\setlength{\tabcolsep}{3.2pt}
\setlength{\aboverulesep}{0.3ex}
\setlength{\belowrulesep}{0.4ex}
\renewcommand{\arraystretch}{1.05}
\begin{tabular}{@{}llccccccc@{}}
\toprule
\multicolumn{9}{c}{\textbf{Fail2Drive-ID}} \\
\midrule
\textbf{Method} & \textbf{Selection metric} & \multicolumn{4}{c}{\textbf{NMAE $\downarrow$}} & \multicolumn{3}{c}{\textbf{SP $\uparrow$}} \\
\cmidrule(lr){3-6}\cmidrule(lr){7-9}
 &  & \textbf{HM} & \textbf{DR} & \textbf{SR} & \textbf{Mean} & \textbf{HM} & \textbf{DR} & \textbf{SR} \\
\midrule
\rowcolor{black!5}
Ours & All metrics & 0.294 & 0.225 & 0.310 & 0.276 & 0.785 & 0.835 & 0.770 \\
Random & None & 0.611 & 0.488 & 0.643 & 0.580 & 0.571 & 0.646 & 0.558 \\
\midrule
\multirow{3}{*}{tinyBenchmarks} & HM & \underline{0.804} & 0.702 & 0.820 & 0.775 & \underline{0.482} & 0.608 & 0.296 \\
 & DR & 0.766 & \underline{0.681} & 0.814 & 0.754 & 0.477 & \underline{0.583} & 0.506 \\
 & \bestcell{SR} & \bestcell{0.785} & \bestcell{0.602} & \bestcell{\underline{0.837}} & \bestcell{0.741} & \bestcell{0.848} & \bestcell{0.883} & \bestcell{\underline{0.653}} \\
\midrule
\multirow{3}{*}{ZipBench} & \bestcell{HM} & \bestcell{\underline{0.544}} & \bestcell{0.396} & \bestcell{0.592} & \bestcell{0.510} & \bestcell{\underline{0.544}} & \bestcell{0.596} & \bestcell{0.565} \\
 & DR & 0.800 & \underline{0.583} & 0.842 & 0.742 & 0.630 & \underline{0.687} & 0.580 \\
 & SR & 0.766 & 0.668 & \underline{0.809} & 0.748 & 0.717 & 0.776 & \underline{0.671} \\
\midrule
\multirow{3}{*}{mRMR} & HM & \underline{0.439} & 0.383 & 0.440 & 0.420 & \underline{0.593} & 0.550 & 0.357 \\
   & \bestcell{DR} & \bestcell{0.403} & \bestcell{\underline{0.296}} & \bestcell{0.405} & \bestcell{0.368} & \bestcell{0.507} & \bestcell{\underline{0.614}} & \bestcell{0.345} \\
  & SR & 0.424 & 0.453 & \underline{0.457} & 0.445 & 0.561 & 0.496 & \underline{0.591} \\
\midrule
\multirow{3}{*}{AnchorPoints} & \bestcell{HM} & \bestcell{\underline{0.290}} & \bestcell{0.235} & \bestcell{0.327} & \bestcell{0.284} & \bestcell{\underline{0.705}} & \bestcell{0.684} & \bestcell{0.726} \\
  & DR & 0.698 & \underline{0.696} & 0.704 & 0.699 & 0.551 & \underline{0.604} & 0.602 \\
  & SR & 0.914 & 0.874 & \underline{0.969} & 0.919 & 0.699 & 0.712 & \underline{0.728} \\
\midrule
\multirow{3}{*}{EssenceBench} & HM & \underline{1.113} & 0.997 & 1.105 & 1.072 & \underline{0.518} & 0.496 & 0.615 \\
 & \bestcell{DR} & \bestcell{1.031} & \bestcell{\underline{0.930}} & \bestcell{1.027} & \bestcell{0.996} & \bestcell{0.432} & \bestcell{\underline{0.550}} & \bestcell{0.551} \\
 & SR & 1.294 & 1.056 & \underline{1.312} & 1.220 & 0.543 & 0.596 & \underline{0.648} \\
\midrule
\multicolumn{9}{c}{\textbf{Fail2Drive-OOD}} \\
\midrule
\textbf{Method} & \textbf{Selection metric} & \multicolumn{4}{c}{\textbf{NMAE $\downarrow$}} & \multicolumn{3}{c}{\textbf{SP $\uparrow$}} \\
\cmidrule(lr){3-6}\cmidrule(lr){7-9}
 &  & \textbf{HM} & \textbf{DR} & \textbf{SR} & \textbf{Mean} & \textbf{HM} & \textbf{DR} & \textbf{SR} \\
\midrule
\rowcolor{black!5}
Ours & All metrics & 0.421 & 0.295 & 0.451 & 0.389 & 0.865 & 0.869 & 0.858 \\
Random & None & 0.922 & 0.501 & 0.958 & 0.794 & 0.777 & 0.865 & 0.728 \\
\midrule
\multirow{3}{*}{tinyBenchmarks} & \bestcell{HM} & \bestcell{\underline{0.687}} & \bestcell{0.637} & \bestcell{0.707} & \bestcell{0.677} & \bestcell{\underline{0.702}} & \bestcell{0.693} & \bestcell{0.664} \\
 & DR & 0.979 & \underline{0.528} & 1.107 & 0.872 & 0.358 & \underline{0.675} & 0.152 \\
 & SR & 0.889 & 0.660 & \underline{0.968} & 0.839 & 0.544 & 0.544 & \underline{0.544} \\
\midrule
\multirow{3}{*}{ZipBench} & HM & \underline{0.930} & 0.766 & 0.973 & 0.889 & \underline{0.635} & 0.683 & 0.573 \\
 & DR & 0.830 & \underline{0.532} & 0.874 & 0.745 & 0.492 & \underline{0.651} & 0.539 \\
 & \bestcell{SR} & \bestcell{0.784} & \bestcell{0.612} & \bestcell{\underline{0.808}} & \bestcell{0.735} & \bestcell{0.635} & \bestcell{0.731} & \bestcell{\underline{0.586}} \\
\midrule
\multirow{3}{*}{mRMR} & HM & \underline{1.853} & 0.389 & 0.590 & 0.944 & \underline{0.129} & 0.539 & 0.675 \\
   & \bestcell{DR} & \bestcell{0.710} & \bestcell{\underline{0.352}} & \bestcell{0.440} & \bestcell{0.501} & \bestcell{0.632} & \bestcell{\underline{0.482}} & \bestcell{0.424} \\
  & SR & 0.657 & 0.611 & \underline{0.709} & 0.659 & 0.407 & 0.396 & \underline{0.369} \\
\midrule
\multirow{3}{*}{AnchorPoints} & HM & \underline{0.710} & 0.465 & 0.716 & 0.630 & \underline{0.816} & 0.799 & 0.753 \\
   & \bestcell{DR} & \bestcell{0.633} & \bestcell{\underline{0.496}} & \bestcell{0.704} & \bestcell{0.611} & \bestcell{0.795} & \bestcell{\underline{0.743}} & \bestcell{0.852} \\
  & SR & 0.916 & 0.633 & \underline{0.987} & 0.845 & 0.710 & 0.711 & \underline{0.706} \\
\midrule
\multirow{3}{*}{EssenceBench} & HM & \underline{0.723} & 0.539 & 0.758 & 0.673 & \underline{0.704} & 0.768 & 0.784 \\
 & DR & 0.723 & \underline{0.539} & 0.758 & 0.673 & 0.704 & \underline{0.768} & 0.784 \\
 & \bestcell{SR} & \bestcell{0.614} & \bestcell{0.476} & \bestcell{\underline{0.652}} & \bestcell{0.581} & \bestcell{0.725} & \bestcell{0.746} & \bestcell{\underline{0.796}} \\
\bottomrule
\end{tabular}
\end{table}

\begin{table}[!htbp]
\setlength{\belowcaptionskip}{8pt}
\centering
\caption{\textbf{Cross-metric fidelity on LiveIdea-Core3 and WorkBench at 20\% retention.} Ours and Random do not use metric-specific selection. The Mean column averages NMAE across metrics; pale blue-gray marks each single-metric baseline\textquotesingle s lowest-mean selection target.}
\label{tab:cross-group-3-language}
\small
\setlength{\tabcolsep}{1.4pt}
\setlength{\aboverulesep}{0.3ex}
\setlength{\belowrulesep}{0.4ex}
\renewcommand{\arraystretch}{1.05}
\resizebox{\linewidth}{!}{%
\begin{tabular}{@{}llccccccc@{}}
\toprule
\multicolumn{9}{c}{\textbf{LiveIdea-Core3}} \\
\midrule
\textbf{Method} & \textbf{Selection metric} & \multicolumn{4}{c}{\textbf{NMAE $\downarrow$}} & \multicolumn{3}{c}{\textbf{SP $\uparrow$}} \\
\cmidrule(lr){3-6}\cmidrule(lr){7-9}
 &  & \textbf{Originality} & \textbf{Feasibility} & \textbf{Clarity} & \textbf{Mean} & \textbf{Originality} & \textbf{Feasibility} & \textbf{Clarity} \\
\midrule
\rowcolor{black!5}
Ours & All metrics & 0.071 & 0.130 & 0.107 & 0.103 & 0.995 & 0.977 & 0.974 \\
Random & None & 0.071 & 0.149 & 0.103 & 0.108 & 0.995 & 0.976 & 0.979 \\
\midrule
\multirow{3}{*}{tinyBenchmarks} & Originality & \underline{0.103} & 0.382 & 0.208 & 0.231 & \underline{0.990} & 0.963 & 0.956 \\
 & \bestcell{Feasibility} & \bestcell{0.117} & \bestcell{\underline{0.187}} & \bestcell{0.184} & \bestcell{0.162} & \bestcell{0.985} & \bestcell{\underline{0.952}} & \bestcell{0.946} \\
 & Clarity & 0.122 & 0.274 & \underline{0.179} & 0.191 & 0.987 & 0.965 & \underline{0.941} \\
\midrule
\multirow{3}{*}{ZipBench} & Originality & \underline{0.114} & 0.228 & 0.171 & 0.171 & \underline{0.988} & 0.949 & 0.954 \\
 & Feasibility & 0.138 & \underline{0.334} & 0.154 & 0.208 & 0.986 & \underline{0.942} & 0.949 \\
 & \bestcell{Clarity} & \bestcell{0.111} & \bestcell{0.228} & \bestcell{\underline{0.155}} & \bestcell{0.165} & \bestcell{0.988} & \bestcell{0.948} & \bestcell{\underline{0.956}} \\
\midrule
\multirow{3}{*}{mRMR} & Originality & \underline{0.136} & 0.442 & 0.499 & 0.359 & \underline{0.994} & 0.946 & 0.950 \\
  & Feasibility & 0.096 & \underline{0.486} & 0.514 & 0.365 & 0.996 & \underline{0.960} & 0.929 \\
   & \bestcell{Clarity} & \bestcell{0.117} & \bestcell{0.414} & \bestcell{\underline{0.474}} & \bestcell{0.335} & \bestcell{0.994} & \bestcell{0.958} & \bestcell{\underline{0.954}} \\
\midrule
\multirow{3}{*}{AnchorPoints} & \bestcell{Originality} & \bestcell{\underline{0.091}} & \bestcell{0.151} & \bestcell{0.120} & \bestcell{0.121} & \bestcell{\underline{0.994}} & \bestcell{0.976} & \bestcell{0.974} \\
  & Feasibility & 0.097 & \underline{0.177} & 0.142 & 0.139 & 0.995 & \underline{0.969} & 0.968 \\
  & Clarity & 0.088 & 0.212 & \underline{0.145} & 0.149 & 0.992 & 0.976 & \underline{0.968} \\
\midrule
\multirow{3}{*}{EssenceBench} & Originality & \underline{0.075} & 0.152 & 0.086 & 0.104 & \underline{0.996} & 0.986 & 0.983 \\
 & Feasibility & 0.059 & \underline{0.161} & 0.085 & 0.102 & 0.997 & \underline{0.983} & 0.976 \\
 & \bestcell{Clarity} & \bestcell{0.073} & \bestcell{0.113} & \bestcell{\underline{0.107}} & \bestcell{0.098} & \bestcell{0.996} & \bestcell{0.988} & \bestcell{\underline{0.967}} \\
\midrule
\multicolumn{9}{c}{\textbf{WorkBench}} \\
\midrule
\textbf{Method} & \textbf{Selection metric} & \multicolumn{4}{c}{\textbf{NMAE $\downarrow$}} & \multicolumn{3}{c}{\textbf{SP $\uparrow$}} \\
\cmidrule(lr){3-6}\cmidrule(lr){7-9}
 &  & \textbf{Correct} & \textbf{Safety} & \textbf{Exact match} & \textbf{Mean} & \textbf{Correct} & \textbf{Safety} & \textbf{Exact match} \\
\midrule
\rowcolor{black!5}
Ours & All metrics & 0.129 & 0.158 & 0.130 & 0.139 & 0.986 & 0.961 & 0.984 \\
Random & None & 0.138 & 0.179 & 0.183 & 0.166 & 0.983 & 0.940 & 0.981 \\
\midrule
\multirow{3}{*}{tinyBenchmarks} & Correct & \underline{0.179} & 0.273 & 0.227 & 0.226 & \underline{0.977} & 0.939 & 0.973 \\
 & \bestcell{Safety} & \bestcell{0.189} & \bestcell{\underline{0.239}} & \bestcell{0.218} & \bestcell{0.215} & \bestcell{0.971} & \bestcell{\underline{0.904}} & \bestcell{0.971} \\
 & Exact match & 0.221 & 0.310 & \underline{0.217} & 0.249 & 0.982 & 0.942 & \underline{0.974} \\
\midrule
\multirow{3}{*}{ZipBench} & \bestcell{Correct} & \bestcell{\underline{0.183}} & \bestcell{0.268} & \bestcell{0.217} & \bestcell{0.223} & \bestcell{\underline{0.986}} & \bestcell{0.954} & \bestcell{0.963} \\
 & Safety & 0.263 & \underline{0.302} & 0.270 & 0.278 & 0.966 & \underline{0.942} & 0.961 \\
 & Exact match & 0.211 & 0.258 & \underline{0.224} & 0.231 & 0.972 & 0.961 & \underline{0.962} \\
\midrule
\multirow{3}{*}{mRMR} & \bestcell{Correct} & \bestcell{\underline{0.323}} & \bestcell{0.765} & \bestcell{0.351} & \bestcell{0.480} & \bestcell{\underline{0.913}} & \bestcell{-0.421} & \bestcell{0.874} \\
  & Safety & 0.323 & \underline{0.765} & 0.351 & 0.480 & 0.913 & \underline{-0.421} & 0.874 \\
  & Exact match & 0.323 & 0.765 & \underline{0.351} & 0.480 & 0.913 & -0.421 & \underline{0.874} \\
\midrule
\multirow{3}{*}{AnchorPoints} & \bestcell{Correct} & \bestcell{\underline{0.749}} & \bestcell{0.940} & \bestcell{0.801} & \bestcell{0.830} & \bestcell{\underline{0.881}} & \bestcell{0.847} & \bestcell{0.807} \\
  & Safety & 1.559 & \underline{2.399} & 1.521 & 1.826 & 0.810 & \underline{0.729} & 0.680 \\
  & Exact match & 0.726 & 0.933 & \underline{0.861} & 0.840 & 0.891 & 0.825 & \underline{0.843} \\
\midrule
\multirow{3}{*}{EssenceBench} & Correct & \underline{0.875} & 1.012 & 0.796 & 0.894 & \underline{0.977} & 0.975 & 0.951 \\
 & Safety & 1.607 & \underline{2.388} & 1.086 & 1.694 & 0.989 & \underline{0.954} & 0.925 \\
 & \bestcell{Exact match} & \bestcell{0.464} & \bestcell{0.321} & \bestcell{\underline{0.344}} & \bestcell{0.376} & \bestcell{0.976} & \bestcell{0.965} & \bestcell{\underline{0.951}} \\
\bottomrule
\end{tabular}
}
\end{table}

\makeatletter
\setlength{\@fptop}{0pt}
\makeatother
\begin{table}[!htbp]
\setlength{\belowcaptionskip}{8pt}
\centering
\caption{\textbf{Cross-metric fidelity on benchmarks with 4 metrics at 20\% retention.} Each baseline selects using the indicated metric and is evaluated on every component. Underlining marks diagonal entries. Ours and Random do not use metric-specific selection. The Mean column averages NMAE across metrics; pale blue-gray marks each single-metric baseline\textquotesingle s lowest-mean selection target.}
\label{tab:cross-group-4}
\scriptsize
\setlength{\tabcolsep}{3.2pt}
\renewcommand{\arraystretch}{1.10}
\resizebox{\linewidth}{!}{%
\begin{tabular}{@{}llccccccccc@{}}
\toprule
\multicolumn{11}{c}{\textbf{Bench2Drive}} \\
\midrule
\textbf{Method} & \textbf{Selection metric} & \multicolumn{5}{c}{\textbf{NMAE $\downarrow$}} & \multicolumn{4}{c}{\textbf{SP $\uparrow$}} \\
\cmidrule(lr){3-7}\cmidrule(lr){8-11}
 &  & \textbf{Efficiency} & \textbf{Score} & \textbf{Smoothness} & \textbf{Success} & \textbf{Mean} & \textbf{Efficiency} & \textbf{Score} & \textbf{Smoothness} & \textbf{Success} \\
\midrule
\rowcolor{black!5}
Ours & All metrics & 0.053 & 0.435 & 0.159 & 0.448 & 0.274 & 0.782 & 0.845 & 0.911 & 0.835 \\
Random & None & 0.139 & 0.631 & 0.253 & 0.529 & 0.388 & 0.733 & 0.857 & 0.933 & 0.844 \\
\midrule
\multirow{4}{*}{tinyBenchmarks} & \bestcell{Efficiency} & \bestcell{\underline{0.077}} & \bestcell{0.714} & \bestcell{0.339} & \bestcell{0.706} & \bestcell{0.459} & \bestcell{\underline{0.766}} & \bestcell{0.770} & \bestcell{0.824} & \bestcell{0.741} \\
 & Score & 0.143 & \underline{0.782} & 0.346 & 0.929 & 0.550 & 0.845 & \underline{0.755} & 0.824 & 0.714 \\
 & Smoothness & 0.123 & 1.176 & \underline{0.255} & 1.174 & 0.682 & 0.658 & 0.806 & \underline{0.836} & 0.656 \\
 & Success & 0.121 & 0.905 & 0.256 & \underline{0.912} & 0.549 & 0.725 & 0.610 & 0.866 & \underline{0.521} \\
\midrule
\multirow{4}{*}{ZipBench} & Efficiency & \underline{0.111} & 0.872 & 0.343 & 0.818 & 0.536 & \underline{0.758} & 0.748 & 0.862 & 0.787 \\
 & \bestcell{Score} & \bestcell{0.100} & \bestcell{\underline{0.831}} & \bestcell{0.338} & \bestcell{0.860} & \bestcell{0.532} & \bestcell{0.668} & \bestcell{\underline{0.750}} & \bestcell{0.869} & \bestcell{0.747} \\
 & Smoothness & 0.174 & 0.944 & \underline{0.299} & 0.973 & 0.598 & 0.847 & 0.783 & \underline{0.845} & 0.825 \\
 & Success & 0.229 & 0.807 & 0.315 & \underline{0.821} & 0.543 & 0.689 & 0.815 & 0.872 & \underline{0.726} \\
\midrule
\multirow{4}{*}{mRMR} & Efficiency & \underline{0.243} & 0.685 & 0.503 & 0.558 & 0.497 & \underline{0.581} & 0.611 & 0.910 & 0.542 \\
   & \bestcell{Score} & \bestcell{0.205} & \bestcell{\underline{0.378}} & \bestcell{0.516} & \bestcell{0.361} & \bestcell{0.365} & \bestcell{0.660} & \bestcell{\underline{0.871}} & \bestcell{0.939} & \bestcell{0.863} \\
  & Smoothness & 0.171 & 0.444 & \underline{0.516} & 0.427 & 0.390 & 0.617 & 0.721 & \underline{0.896} & 0.840 \\
  & Success & 0.271 & 0.501 & 0.632 & \underline{0.440} & 0.461 & 0.553 & 0.786 & 0.863 & \underline{0.763} \\
\midrule
\multirow{4}{*}{AnchorPoints} & \bestcell{Efficiency} & \bestcell{\underline{0.075}} & \bestcell{0.816} & \bestcell{0.523} & \bestcell{0.951} & \bestcell{0.591} & \bestcell{\underline{0.647}} & \bestcell{0.899} & \bestcell{0.788} & \bestcell{0.826} \\
  & Score & 0.446 & \underline{1.913} & 0.734 & 2.066 & 1.290 & 0.672 & \underline{0.631} & 0.849 & 0.641 \\
  & Smoothness & 0.202 & 1.150 & \underline{0.343} & 1.027 & 0.680 & 0.775 & 0.731 & \underline{0.741} & 0.809 \\
  & Success & 0.509 & 2.046 & 0.954 & \underline{2.477} & 1.496 & 0.591 & 0.644 & 0.801 & \underline{0.660} \\
\midrule
\multirow{4}{*}{EssenceBench} & \bestcell{Efficiency} & \bestcell{\underline{0.068}} & \bestcell{0.610} & \bestcell{0.110} & \bestcell{0.559} & \bestcell{0.337} & \bestcell{\underline{0.685}} & \bestcell{0.918} & \bestcell{0.921} & \bestcell{0.892} \\
 & Score & 0.107 & \underline{1.190} & 0.403 & 1.478 & 0.795 & 0.810 & \underline{0.921} & 0.957 & 0.910 \\
 & Smoothness & 0.070 & 0.674 & \underline{0.157} & 0.674 & 0.394 & 0.617 & 0.846 & \underline{0.817} & 0.918 \\
 & Success & 0.134 & 0.863 & 0.160 & \underline{1.195} & 0.588 & 0.821 & 0.907 & 0.975 & \underline{0.920} \\
\midrule
\multicolumn{11}{c}{\textbf{LiveIdea-Full4}} \\
\midrule
\textbf{Method} & \textbf{Selection metric} & \multicolumn{5}{c}{\textbf{NMAE $\downarrow$}} & \multicolumn{4}{c}{\textbf{SP $\uparrow$}} \\
\cmidrule(lr){3-7}\cmidrule(lr){8-11}
 &  & \textbf{Originality} & \textbf{Feasibility} & \textbf{Clarity} & \textbf{Fluency} & \textbf{Mean} & \textbf{Originality} & \textbf{Feasibility} & \textbf{Clarity} & \textbf{Fluency} \\
\midrule
\rowcolor{black!5}
Ours & All metrics & 0.142 & 0.234 & 0.178 & 0.345 & 0.225 & 0.990 & 0.952 & 0.944 & 0.906 \\
Random & None & 0.135 & 0.241 & 0.227 & 0.552 & 0.289 & 0.988 & 0.949 & 0.923 & 0.883 \\
\midrule
\multirow{4}{*}{tinyBenchmarks} & Originality & \underline{0.222} & 0.440 & 0.348 & 0.670 & 0.420 & \underline{0.965} & 0.869 & 0.870 & 0.741 \\
 & \bestcell{Feasibility} & \bestcell{0.190} & \bestcell{\underline{0.371}} & \bestcell{0.354} & \bestcell{0.631} & \bestcell{0.387} & \bestcell{0.969} & \bestcell{\underline{0.861}} & \bestcell{0.899} & \bestcell{0.770} \\
 & Clarity & 0.192 & 0.494 & \underline{0.319} & 0.585 & 0.398 & 0.972 & 0.859 & \underline{0.896} & 0.799 \\
 & Fluency & 0.226 & 0.461 & 0.386 & \underline{0.548} & 0.405 & 0.966 & 0.880 & 0.876 & \underline{0.770} \\
\midrule
\multirow{4}{*}{ZipBench} & Originality & \underline{0.164} & 0.429 & 0.356 & 0.687 & 0.409 & \underline{0.975} & 0.888 & 0.894 & 0.827 \\
 & \bestcell{Feasibility} & \bestcell{0.229} & \bestcell{\underline{0.369}} & \bestcell{0.323} & \bestcell{0.669} & \bestcell{0.398} & \bestcell{0.967} & \bestcell{\underline{0.890}} & \bestcell{0.901} & \bestcell{0.776} \\
 & Clarity & 0.231 & 0.470 & \underline{0.330} & 0.576 & 0.402 & 0.966 & 0.896 & \underline{0.895} & 0.817 \\
 & Fluency & 0.240 & 0.415 & 0.334 & \underline{0.609} & 0.400 & 0.970 & 0.891 & 0.878 & \underline{0.801} \\
\midrule
\multirow{4}{*}{mRMR} & \bestcell{Originality} & \bestcell{\underline{0.181}} & \bestcell{0.418} & \bestcell{0.539} & \bestcell{0.527} & \bestcell{0.416} & \bestcell{\underline{0.981}} & \bestcell{0.911} & \bestcell{0.865} & \bestcell{0.793} \\
  & Feasibility & 0.163 & \underline{0.489} & 0.591 & 0.434 & 0.419 & 0.987 & \underline{0.848} & 0.705 & 0.873 \\
  & Clarity & 0.189 & 0.495 & \underline{0.573} & 0.514 & 0.443 & 0.987 & 0.846 & \underline{0.787} & 0.887 \\
  & Fluency & 0.168 & 0.449 & 0.576 & \underline{0.577} & 0.442 & 0.979 & 0.918 & 0.830 & \underline{0.773} \\
\midrule
\multirow{4}{*}{AnchorPoints} & Originality & \underline{0.169} & 0.306 & 0.248 & 0.476 & 0.300 & \underline{0.974} & 0.923 & 0.947 & 0.888 \\
  & Feasibility & 0.193 & \underline{0.318} & 0.275 & 0.377 & 0.290 & 0.985 & \underline{0.942} & 0.947 & 0.916 \\
   & \bestcell{Clarity} & \bestcell{0.129} & \bestcell{0.261} & \bestcell{\underline{0.229}} & \bestcell{0.459} & \bestcell{0.269} & \bestcell{0.989} & \bestcell{0.947} & \bestcell{\underline{0.938}} & \bestcell{0.889} \\
  & Fluency & 0.141 & 0.343 & 0.257 & \underline{0.544} & 0.321 & 0.982 & 0.928 & 0.932 & \underline{0.847} \\
\midrule
\multirow{4}{*}{EssenceBench} & Originality & \underline{0.173} & 0.284 & 0.213 & 0.386 & 0.264 & \underline{0.993} & 0.920 & 0.934 & 0.922 \\
 & \bestcell{Feasibility} & \bestcell{0.146} & \bestcell{\underline{0.230}} & \bestcell{0.211} & \bestcell{0.426} & \bestcell{0.253} & \bestcell{0.988} & \bestcell{\underline{0.945}} & \bestcell{0.947} & \bestcell{0.896} \\
 & Clarity & 0.159 & 0.227 & \underline{0.204} & 0.426 & 0.254 & 0.985 & 0.951 & \underline{0.955} & 0.867 \\
 & Fluency & 0.133 & 0.477 & 0.220 & \underline{0.447} & 0.319 & 0.987 & 0.933 & 0.952 & \underline{0.909} \\
\bottomrule
\end{tabular}
}
\end{table}

\label{app:additional_cross_metric_results}

The following tables report the remaining benchmark-level cross-metric results at 20\% retention. Entries are NMAE / SP, averaged over three seeds.

\begin{table}[!htbp]
\centering
\caption{\textbf{Cross-metric fidelity on RouterArena at 20\% retention.} Each cell reports NMAE / SP. Underlining marks the selection metric. The Mean column averages NMAE across metrics; pale blue-gray marks each single-metric baseline's lowest-mean selection target.}
\label{tab:cross-additional-routerarena-multi}
\small
\setlength{\tabcolsep}{5.5pt}
\renewcommand{\arraystretch}{1.05}
\begin{tabular}{@{}llccc@{}}
\toprule
Method & Selection metric & \textbf{Acc.} & \textbf{Cost} & \textbf{Mean} \\
\midrule
\rowcolor{black!5}
Ours & All metrics & 0.11/0.96 & 0.07/0.98 & 0.09 \\
\midrule
Random & None & 0.19/1.00 & 0.10/0.98 & 0.15 \\
\midrule
tinyBenchmarks & \bestcell{Acc.} & \bestcell{\underline{0.43/0.89}} & \bestcell{0.14/0.94} & \bestcell{0.28} \\
 & Cost & 0.55/0.90 & \underline{0.16/0.98} & 0.35 \\
\midrule
ZipBench & \bestcell{Acc.} & \bestcell{\underline{0.37/0.96}} & \bestcell{0.12/0.96} & \bestcell{0.25} \\
 & Cost & 0.43/0.92 & \underline{0.11/0.98} & 0.27 \\
\midrule
mRMR & \bestcell{Acc.} & \bestcell{\underline{0.68/0.83}} & \bestcell{1.36/0.89} & \bestcell{1.02} \\
 & Cost & 0.62/0.77 & \underline{1.48/0.77} & 1.05 \\
\midrule
AnchorPoints & Acc. & \underline{9.44/0.46} & 1.13/0.81 & 5.29 \\
  & \bestcell{Cost} & \bestcell{0.43/0.95} & \bestcell{\underline{0.13/0.99}} & \bestcell{0.28} \\
\midrule
EssenceBench & Acc. & \underline{6.39/1.00} & 0.48/0.94 & 3.43 \\
 & \bestcell{Cost} & \bestcell{0.45/0.92} & \bestcell{\underline{0.38/0.94}} & \bestcell{0.42} \\
\bottomrule
\end{tabular}
\end{table}

\begin{table}[p]
\centering
\caption{\textbf{Cross-metric fidelity on ConStory at 20\% retention.} Each cell reports NMAE / SP. Underlining marks the selection metric. The Mean column averages NMAE across metrics; pale blue-gray marks each single-metric baseline's lowest-mean selection target.}
\label{tab:cross-additional-constory}
\scriptsize
\setlength{\tabcolsep}{2.1pt}
\renewcommand{\arraystretch}{1.00}
\resizebox{\textwidth}{!}{%
\begin{tabular}{@{}llcccccc@{}}
\toprule
Method & Selection metric & \textbf{Char.} & \textbf{Fact.} & \textbf{Narr.} & \textbf{Time.} & \textbf{World} & \textbf{Mean} \\
\midrule
\rowcolor{black!5}
Ours & All metrics & 0.51/0.74 & 0.20/0.97 & 0.27/0.79 & 0.23/0.96 & 0.28/0.92 & 0.30 \\
\midrule
Random & None & 0.46/0.79 & 0.23/0.96 & 0.24/0.79 & 0.19/0.97 & 0.36/0.90 & 0.29 \\
\midrule
tinyBenchmarks & Char. & \underline{0.75/0.67} & 0.31/0.93 & 0.50/0.71 & 0.27/0.93 & 0.55/0.71 & 0.48 \\
 & Fact. & 0.89/0.72 & \underline{0.44/0.87} & 0.60/0.77 & 0.31/0.92 & 0.71/0.74 & 0.59 \\
  & \bestcell{Narr.} & \bestcell{0.71/0.74} & \bestcell{0.33/0.87} & \bestcell{\underline{0.38/0.77}} & \bestcell{0.25/0.96} & \bestcell{0.50/0.85} & \bestcell{0.43} \\
 & Time. & 1.04/0.64 & 0.38/0.88 & 0.43/0.78 & \underline{0.34/0.91} & 0.62/0.79 & 0.56 \\
 & World & 0.56/0.79 & 0.38/0.91 & 0.43/0.76 & 0.29/0.90 & \underline{0.62/0.80} & 0.45 \\
\bottomrule
\end{tabular}%
}
\end{table}

\begin{table}[p]
\centering
\caption{\textbf{Cross-metric fidelity on SoB-Audio at 20\% retention.} Each cell reports NMAE / SP. Underlining marks the selection metric. The Mean column averages NMAE across metrics; pale blue-gray marks each single-metric baseline's lowest-mean selection target.}
\label{tab:cross-additional-sob-audio}
\scriptsize
\setlength{\tabcolsep}{2.1pt}
\renewcommand{\arraystretch}{1.00}
\resizebox{\textwidth}{!}{%
\begin{tabular}{@{}llcccccc@{}}
\toprule
Method & Selection metric & \textbf{Schema} & \textbf{LeafEM} & \textbf{ValueF1} & \textbf{PathRec} & \textbf{PathF1} & \textbf{Mean} \\
\midrule
\rowcolor{black!5}
Ours & All metrics & 0.55/0.77 & 0.73/0.83 & 0.54/0.92 & 0.49/0.84 & 0.58/0.79 & 0.58 \\
\midrule
Random & None & 0.56/0.73 & 1.14/0.75 & 0.76/0.86 & 0.50/0.82 & 0.57/0.73 & 0.70 \\
\midrule
tinyBenchmarks & Schema & \underline{0.82/0.70} & 2.02/0.65 & 1.45/0.75 & 0.90/0.73 & 0.96/0.63 & 1.23 \\
 & \bestcell{LeafEM} & \bestcell{0.89/0.64} & \bestcell{\underline{1.18/0.59}} & \bestcell{0.93/0.73} & \bestcell{0.98/0.69} & \bestcell{0.97/0.58} & \bestcell{0.99} \\
 & ValueF1 & 0.88/0.66 & 1.53/0.57 & \underline{0.97/0.79} & 0.76/0.67 & 0.94/0.55 & 1.02 \\
 & PathRec & 0.71/0.74 & 2.37/0.73 & 1.68/0.78 & \underline{0.75/0.73} & 0.89/0.70 & 1.28 \\
 & PathF1 & 0.89/0.58 & 1.60/0.50 & 0.91/0.62 & 0.84/0.51 & \underline{0.95/0.49} & 1.04 \\
\bottomrule
\end{tabular}%
}
\end{table}

\begin{table}[p]
\centering
\caption{\textbf{Cross-metric fidelity on SoB-Image at 20\% retention.} Each cell reports NMAE / SP. Underlining marks the selection metric. The Mean column averages NMAE across metrics; pale blue-gray marks each single-metric baseline's lowest-mean selection target.}
\label{tab:cross-additional-sob-image}
\scriptsize
\setlength{\tabcolsep}{2.1pt}
\renewcommand{\arraystretch}{1.00}
\resizebox{\textwidth}{!}{%
\begin{tabular}{@{}llcccccc@{}}
\toprule
Method & Selection metric & \textbf{PathRec} & \textbf{LeafEM} & \textbf{PathF1} & \textbf{Schema} & \textbf{ValueF1} & \textbf{Mean} \\
\midrule
\rowcolor{black!5}
Ours & All metrics & 0.51/0.77 & 0.38/0.88 & 0.56/0.74 & 0.54/0.80 & 0.58/0.67 & 0.51 \\
\midrule
Random & None & 1.01/0.68 & 1.20/0.77 & 1.10/0.57 & 1.00/0.62 & 1.57/0.62 & 1.17 \\
\midrule
tinyBenchmarks & PathRec & \underline{0.95/0.60} & 1.37/0.69 & 1.11/0.54 & 0.95/0.63 & 1.90/0.52 & 1.25 \\
 & \bestcell{LeafEM} & \bestcell{0.87/0.64} & \bestcell{\underline{1.19/0.71}} & \bestcell{0.94/0.51} & \bestcell{0.93/0.59} & \bestcell{1.20/0.64} & \bestcell{1.03} \\
 & PathF1 & 1.16/0.62 & 1.65/0.70 & \underline{1.30/0.52} & 1.10/0.62 & 1.98/0.59 & 1.44 \\
 & Schema & 1.06/0.60 & 1.15/0.69 & 1.21/0.48 & \underline{1.02/0.55} & 1.37/0.32 & 1.16 \\
 & ValueF1 & 1.14/0.64 & 1.22/0.65 & 1.28/0.52 & 1.14/0.61 & \underline{1.50/0.55} & 1.26 \\
\bottomrule
\end{tabular}%
}
\end{table}

\begin{table}[p]
\centering
\caption{\textbf{Cross-metric fidelity on SoB-Text at 20\% retention.} Each cell reports NMAE / SP. Underlining marks the selection metric. The Mean column averages NMAE across metrics; pale blue-gray marks each single-metric baseline's lowest-mean selection target.}
\label{tab:cross-additional-sob-text}
\scriptsize
\setlength{\tabcolsep}{2.1pt}
\renewcommand{\arraystretch}{1.00}
\resizebox{\textwidth}{!}{%
\begin{tabular}{@{}llccccccc@{}}
\toprule
Method & Selection metric & \textbf{Schema} & \textbf{LeafEM} & \textbf{ValueF1} & \textbf{PathRec} & \textbf{PathF1} & \textbf{JSONEM} & \textbf{Mean} \\
\midrule
\rowcolor{black!5}
Ours & All metrics & 0.14/0.97 & 0.16/0.95 & 0.18/0.96 & 0.14/0.97 & 0.14/0.97 & 0.17/0.97 & 0.16 \\
\midrule
Random & None & 0.19/0.95 & 0.21/0.97 & 0.19/0.96 & 0.18/0.96 & 0.18/0.95 & 0.32/0.96 & 0.21 \\
\midrule
tinyBenchmarks & Schema & \underline{0.36/0.91} & 0.45/0.83 & 0.44/0.83 & 0.36/0.87 & 0.39/0.89 & 0.65/0.90 & 0.44 \\
 & LeafEM & 0.37/0.91 & \underline{0.34/0.90} & 0.35/0.88 & 0.36/0.89 & 0.35/0.89 & 0.70/0.91 & 0.41 \\
 & \bestcell{ValueF1} & \bestcell{0.31/0.92} & \bestcell{0.45/0.79} & \bestcell{\underline{0.46/0.79}} & \bestcell{0.31/0.93} & \bestcell{0.31/0.90} & \bestcell{0.50/0.87} & \bestcell{0.39} \\
 & PathRec & 0.35/0.84 & 0.53/0.86 & 0.50/0.83 & \underline{0.38/0.81} & 0.37/0.78 & 0.56/0.88 & 0.45 \\
 & PathF1 & 0.39/0.81 & 0.71/0.85 & 0.71/0.87 & 0.38/0.81 & \underline{0.43/0.81} & 0.51/0.86 & 0.52 \\
\bottomrule
\end{tabular}%
}
\end{table}

\begin{table}[p]
\centering
\caption{\textbf{Cross-metric fidelity on MMDocRAG-MM at 20\% retention.} Each cell reports NMAE / SP. Underlining marks the selection metric. The Mean column averages NMAE across metrics; pale blue-gray marks each single-metric baseline's lowest-mean selection target.}
\label{tab:cross-additional-mmdocrag-multimodal}
\scriptsize
\setlength{\tabcolsep}{2.1pt}
\renewcommand{\arraystretch}{1.00}
\resizebox{\textwidth}{!}{%
\begin{tabular}{@{}llccccccccc@{}}
\toprule
Method & Selection metric & \rotatebox{55}{\textbf{QuoteP}} & \rotatebox{55}{\textbf{QuoteR}} & \rotatebox{55}{\textbf{QuoteF1}} & \rotatebox{55}{\textbf{Fluency}} & \rotatebox{55}{\textbf{CitQual}} & \rotatebox{55}{\textbf{TxtImg}} & \rotatebox{55}{\textbf{Logic}} & \rotatebox{55}{\textbf{Fact.}} & \textbf{Mean} \\
\midrule
\rowcolor{black!5}
Ours & All metrics & 0.09/0.99 & 0.05/0.99 & 0.07/0.99 & 0.06/1.00 & 0.06/0.99 & 0.06/0.99 & 0.07/1.00 & 0.06/0.99 & 0.06 \\
\midrule
Random & None & 0.11/0.99 & 0.06/1.00 & 0.07/0.99 & 0.07/0.99 & 0.06/0.99 & 0.05/0.99 & 0.08/0.99 & 0.09/0.99 & 0.07 \\
\midrule
tinyBenchmarks & \bestcell{QuoteP} & \bestcell{\underline{0.19/0.97}} & \bestcell{0.12/0.99} & \bestcell{0.15/0.98} & \bestcell{0.12/0.97} & \bestcell{0.12/0.98} & \bestcell{0.11/0.98} & \bestcell{0.15/0.97} & \bestcell{0.13/0.98} & \bestcell{0.14} \\
 & QuoteR & 0.21/0.95 & \underline{0.13/0.98} & 0.16/0.98 & 0.14/0.98 & 0.15/0.97 & 0.15/0.98 & 0.13/0.98 & 0.14/0.97 & 0.15 \\
 & QuoteF1 & 0.15/0.98 & 0.10/0.99 & \underline{0.11/0.98} & 0.13/0.98 & 0.14/0.98 & 0.12/0.98 & 0.18/0.97 & 0.19/0.98 & 0.14 \\
 & Fluency & 0.27/0.93 & 0.15/0.98 & 0.18/0.97 & \underline{0.16/0.96} & 0.17/0.97 & 0.14/0.98 & 0.20/0.98 & 0.16/0.98 & 0.18 \\
 & CitQual & 0.23/0.96 & 0.13/0.99 & 0.17/0.98 & 0.12/0.98 & \underline{0.12/0.98} & 0.11/0.98 & 0.15/0.98 & 0.16/0.98 & 0.15 \\
\bottomrule
\end{tabular}%
}
\end{table}

\begin{table}[p]
\centering
\caption{\textbf{Cross-metric fidelity on MMDocRAG-Text at 20\% retention.} Each cell reports NMAE / SP. Underlining marks the selection metric. The Mean column averages NMAE across metrics; pale blue-gray marks each single-metric baseline's lowest-mean selection target.}
\label{tab:cross-additional-mmdocrag-pure-text}
\scriptsize
\setlength{\tabcolsep}{2.1pt}
\renewcommand{\arraystretch}{1.00}
\resizebox{\textwidth}{!}{%
\begin{tabular}{@{}llccccccccc@{}}
\toprule
Method & Selection metric & \rotatebox{55}{\textbf{QuoteP}} & \rotatebox{55}{\textbf{QuoteR}} & \rotatebox{55}{\textbf{QuoteF1}} & \rotatebox{55}{\textbf{Fluency}} & \rotatebox{55}{\textbf{CitQual}} & \rotatebox{55}{\textbf{TxtImg}} & \rotatebox{55}{\textbf{Logic}} & \rotatebox{55}{\textbf{Fact.}} & \textbf{Mean} \\
\midrule
\rowcolor{black!5}
Ours & All metrics & 0.09/0.99 & 0.08/0.99 & 0.10/0.98 & 0.08/0.98 & 0.10/0.98 & 0.10/0.98 & 0.11/0.98 & 0.12/0.97 & 0.10 \\
\midrule
Random & None & 0.12/0.99 & 0.09/0.99 & 0.13/0.98 & 0.09/0.98 & 0.11/0.98 & 0.11/0.97 & 0.12/0.98 & 0.12/0.98 & 0.11 \\
\midrule
tinyBenchmarks & QuoteP & \underline{0.13/0.98} & 0.13/0.98 & 0.15/0.95 & 0.13/0.97 & 0.16/0.96 & 0.15/0.96 & 0.14/0.97 & 0.17/0.95 & 0.14 \\
 & QuoteR & 0.16/0.97 & \underline{0.15/0.98} & 0.19/0.95 & 0.16/0.97 & 0.19/0.96 & 0.17/0.96 & 0.21/0.95 & 0.29/0.94 & 0.19 \\
 & \bestcell{QuoteF1} & \bestcell{0.13/0.98} & \bestcell{0.12/0.98} & \bestcell{\underline{0.14/0.97}} & \bestcell{0.12/0.97} & \bestcell{0.14/0.97} & \bestcell{0.14/0.96} & \bestcell{0.14/0.98} & \bestcell{0.20/0.95} & \bestcell{0.14} \\
 & Fluency & 0.19/0.98 & 0.16/0.98 & 0.24/0.96 & \underline{0.15/0.96} & 0.17/0.97 & 0.19/0.96 & 0.23/0.94 & 0.30/0.92 & 0.20 \\
 & CitQual & 0.15/0.97 & 0.16/0.97 & 0.21/0.96 & 0.15/0.96 & \underline{0.16/0.95} & 0.16/0.95 & 0.17/0.97 & 0.23/0.94 & 0.17 \\
\bottomrule
\end{tabular}%
}
\end{table}

\begin{table}[p]
\centering
\caption{\textbf{Cross-metric NMAE on Widget2Code at 20\% retention.} Underlining marks the selection metric. The Mean column averages NMAE across metrics; pale blue-gray marks each single-metric baseline's lowest-mean selection target.}
\label{tab:cross-additional-widget2code}
\scriptsize
\setlength{\tabcolsep}{2.1pt}
\renewcommand{\arraystretch}{1.00}
\resizebox*{!}{0.88\textheight}{%
\begin{tabular}{@{}llccccccccccccc@{}}
\toprule
Method & Selection metric & \rotatebox{55}{\textbf{Margin}} & \rotatebox{55}{\textbf{Content}} & \rotatebox{55}{\textbf{Area}} & \rotatebox{55}{\textbf{TextJac}} & \rotatebox{55}{\textbf{Contrast}} & \rotatebox{55}{\textbf{Local}} & \rotatebox{55}{\textbf{Palette}} & \rotatebox{55}{\textbf{Vibrancy}} & \rotatebox{55}{\textbf{Polarity}} & \rotatebox{55}{\textbf{SSIM}} & \rotatebox{55}{\textbf{LP}} & \rotatebox{55}{\textbf{Geom.}} & \textbf{Mean} \\
\midrule
\rowcolor{black!5}
Ours & All metrics & 0.27 & 0.10 & 0.10 & 0.10 & 0.51 & 0.24 & 0.32 & 0.30 & 0.33 & 0.19 & 0.10 & 0.04 & 0.22 \\
\midrule
Random & None & 0.35 & 0.23 & 0.13 & 0.12 & 0.56 & 0.32 & 0.82 & 0.37 & 0.29 & 0.21 & 0.13 & 0.05 & 0.30 \\
\midrule
tinyBenchmarks & Margin & \underline{0.49} & 0.24 & 0.23 & 0.18 & 0.86 & 0.47 & 1.08 & 0.68 & 1.07 & 0.50 & 0.21 & 0.07 & 0.51 \\
 & \bestcell{Content} & \bestcell{0.49} & \bestcell{\underline{0.20}} & \bestcell{0.29} & \bestcell{0.19} & \bestcell{0.95} & \bestcell{0.32} & \bestcell{1.14} & \bestcell{0.40} & \bestcell{0.47} & \bestcell{0.36} & \bestcell{0.20} & \bestcell{0.08} & \bestcell{0.42} \\
 & Area & 0.53 & 0.31 & \underline{0.27} & 0.18 & 0.92 & 0.47 & 1.07 & 0.48 & 0.57 & 0.38 & 0.24 & 0.08 & 0.46 \\
 & TextJac & 0.51 & 0.28 & 0.21 & \underline{0.14} & 1.05 & 0.40 & 0.90 & 0.88 & 0.71 & 0.26 & 0.25 & 0.09 & 0.47 \\
 & Contrast & 0.47 & 0.25 & 0.40 & 0.24 & \underline{0.79} & 0.56 & 0.89 & 1.00 & 0.61 & 0.36 & 0.29 & 0.11 & 0.50 \\
\bottomrule
\end{tabular}%
}
\end{table}
\begin{table}[p]
\centering
\caption{\textbf{Cross-metric SP on Widget2Code at 20\% retention.} Underlining marks the selection metric; pale blue-gray marks each single-metric baseline's selection target with the lowest mean NMAE.}
\label{tab:cross-additional-widget2code-sp}
\scriptsize
\setlength{\tabcolsep}{2.1pt}
\renewcommand{\arraystretch}{1.00}
\resizebox*{!}{0.88\textheight}{%
\begin{tabular}{@{}llcccccccccccc@{}}
\toprule
Method & Selection metric & \rotatebox{55}{\textbf{Margin}} & \rotatebox{55}{\textbf{Content}} & \rotatebox{55}{\textbf{Area}} & \rotatebox{55}{\textbf{TextJac}} & \rotatebox{55}{\textbf{Contrast}} & \rotatebox{55}{\textbf{Local}} & \rotatebox{55}{\textbf{Palette}} & \rotatebox{55}{\textbf{Vibrancy}} & \rotatebox{55}{\textbf{Polarity}} & \rotatebox{55}{\textbf{SSIM}} & \rotatebox{55}{\textbf{LP}} & \rotatebox{55}{\textbf{Geom.}} \\
\midrule
\rowcolor{black!5}
Ours & All metrics & 0.90 & 0.98 & 0.99 & 0.95 & 0.86 & 0.94 & 0.92 & 0.93 & 0.99 & 0.94 & 0.96 & 0.98 \\
\midrule
Random & None & 0.92 & 0.99 & 0.95 & 0.93 & 0.87 & 0.98 & 0.83 & 0.88 & 0.99 & 0.98 & 0.93 & 0.96 \\
\midrule
tinyBenchmarks & Margin & \underline{0.71} & 0.96 & 0.99 & 0.86 & 0.79 & 0.89 & 0.89 & 0.93 & 0.87 & 0.92 & 0.95 & 0.88 \\
 & \bestcell{Content} & \bestcell{0.88} & \bestcell{\underline{0.93}} & \bestcell{0.95} & \bestcell{0.85} & \bestcell{0.75} & \bestcell{0.94} & \bestcell{0.61} & \bestcell{0.89} & \bestcell{0.93} & \bestcell{0.96} & \bestcell{0.95} & \bestcell{0.94} \\
 & Area & 0.86 & 0.82 & \underline{0.89} & 0.89 & 0.73 & 0.88 & 0.80 & 0.87 & 0.90 & 0.92 & 0.95 & 0.87 \\
 & TextJac & 0.88 & 0.85 & 0.93 & \underline{0.92} & 0.86 & 0.85 & 0.73 & 0.76 & 0.90 & 0.94 & 0.96 & 0.90 \\
 & Contrast & 0.80 & 0.92 & 0.86 & 0.92 & \underline{0.70} & 0.89 & 0.90 & 0.88 & 0.87 & 0.86 & 0.94 & 0.88 \\
\bottomrule
\end{tabular}%
}
\end{table}

\begin{table}[p]
\centering
\caption{\textbf{Cross-metric NMAE on LongForm at 20\% retention.} Underlining marks the selection metric. The Mean column averages NMAE across metrics; pale blue-gray marks each single-metric baseline's lowest-mean selection target.}
\label{tab:cross-additional-longform}
\scriptsize
\setlength{\tabcolsep}{2.1pt}
\renewcommand{\arraystretch}{1.00}
\resizebox*{!}{0.88\textheight}{%
\begin{tabular}{@{}llccccccccccccccc@{}}
\toprule
Method & Selection metric & \rotatebox{55}{\textbf{Amat.}} & \rotatebox{55}{\textbf{Profile}} & \rotatebox{55}{\textbf{Coher.}} & \rotatebox{55}{\textbf{Plot}} & \rotatebox{55}{\textbf{Engage}} & \rotatebox{55}{\textbf{Prompt}} & \rotatebox{55}{\textbf{Plan}} & \rotatebox{55}{\textbf{Poetry}} & \rotatebox{55}{\textbf{Nuance}} & \rotatebox{55}{\textbf{Purple}} & \rotatebox{55}{\textbf{Tell}} & \rotatebox{55}{\textbf{Uncreat.}} & \rotatebox{55}{\textbf{Dialogue}} & \rotatebox{55}{\textbf{Tone}} & \textbf{Mean} \\
\midrule
\rowcolor{black!5}
Ours & All metrics & 0.11 & 0.11 & 0.12 & 0.11 & 0.10 & 0.12 & 0.11 & 0.16 & 0.10 & 0.14 & 0.11 & 0.13 & 0.11 & 0.10 & 0.12 \\
\midrule
Random & None & 0.12 & 0.13 & 0.14 & 0.13 & 0.13 & 0.14 & 0.15 & 0.20 & 0.13 & 0.18 & 0.12 & 0.13 & 0.13 & 0.13 & 0.14 \\
\midrule
tinyBenchmarks & Amat. & \underline{0.17} & 0.18 & 0.22 & 0.22 & 0.16 & 0.23 & 0.20 & 0.27 & 0.16 & 0.24 & 0.17 & 0.22 & 0.17 & 0.18 & 0.20 \\
 & Profile & 0.17 & \underline{0.19} & 0.19 & 0.20 & 0.18 & 0.19 & 0.21 & 0.26 & 0.17 & 0.24 & 0.18 & 0.17 & 0.18 & 0.20 & 0.19 \\
 & Coher. & 0.19 & 0.17 & \underline{0.23} & 0.20 & 0.18 & 0.20 & 0.22 & 0.33 & 0.18 & 0.35 & 0.18 & 0.22 & 0.19 & 0.19 & 0.22 \\
 & Plot & 0.18 & 0.16 & 0.17 & \underline{0.17} & 0.16 & 0.18 & 0.18 & 0.21 & 0.16 & 0.22 & 0.17 & 0.18 & 0.17 & 0.17 & 0.18 \\
 & Engage & 0.14 & 0.18 & 0.17 & 0.18 & \underline{0.16} & 0.20 & 0.19 & 0.21 & 0.16 & 0.20 & 0.16 & 0.16 & 0.17 & 0.16 & 0.17 \\
\bottomrule
\end{tabular}%
}
\end{table}
\begin{table}[p]
\centering
\caption{\textbf{Cross-metric SP on LongForm at 20\% retention.} Underlining marks the selection metric; pale blue-gray marks each single-metric baseline's selection target with the lowest mean NMAE.}
\label{tab:cross-additional-longform-sp}
\scriptsize
\setlength{\tabcolsep}{2.1pt}
\renewcommand{\arraystretch}{1.00}
\resizebox*{!}{0.88\textheight}{%
\begin{tabular}{@{}llcccccccccccccc@{}}
\toprule
Method & Selection metric & \rotatebox{55}{\textbf{Amat.}} & \rotatebox{55}{\textbf{Profile}} & \rotatebox{55}{\textbf{Coher.}} & \rotatebox{55}{\textbf{Plot}} & \rotatebox{55}{\textbf{Engage}} & \rotatebox{55}{\textbf{Prompt}} & \rotatebox{55}{\textbf{Plan}} & \rotatebox{55}{\textbf{Poetry}} & \rotatebox{55}{\textbf{Nuance}} & \rotatebox{55}{\textbf{Purple}} & \rotatebox{55}{\textbf{Tell}} & \rotatebox{55}{\textbf{Uncreat.}} & \rotatebox{55}{\textbf{Dialogue}} & \rotatebox{55}{\textbf{Tone}} \\
\midrule
\rowcolor{black!5}
Ours & All metrics & 0.99 & 0.99 & 0.97 & 0.98 & 0.99 & 0.98 & 0.98 & 0.98 & 0.99 & 0.98 & 0.99 & 0.99 & 0.99 & 0.99 \\
\midrule
Random & None & 0.99 & 0.98 & 0.97 & 0.97 & 0.98 & 0.98 & 0.98 & 0.97 & 0.98 & 0.98 & 0.99 & 0.98 & 0.98 & 0.98 \\
\midrule
tinyBenchmarks & Amat. & \underline{0.98} & 0.97 & 0.94 & 0.96 & 0.97 & 0.97 & 0.97 & 0.93 & 0.97 & 0.96 & 0.98 & 0.97 & 0.97 & 0.97 \\
 & Profile & 0.98 & \underline{0.96} & 0.95 & 0.97 & 0.97 & 0.97 & 0.97 & 0.91 & 0.98 & 0.95 & 0.98 & 0.97 & 0.97 & 0.97 \\
 & Coher. & 0.95 & 0.97 & \underline{0.93} & 0.95 & 0.97 & 0.95 & 0.97 & 0.95 & 0.97 & 0.95 & 0.97 & 0.96 & 0.96 & 0.97 \\
 & Plot & 0.98 & 0.97 & 0.97 & \underline{0.96} & 0.98 & 0.97 & 0.97 & 0.95 & 0.98 & 0.96 & 0.97 & 0.98 & 0.97 & 0.97 \\
 & Engage & 0.97 & 0.96 & 0.96 & 0.96 & \underline{0.98} & 0.97 & 0.98 & 0.94 & 0.97 & 0.95 & 0.97 & 0.98 & 0.97 & 0.97 \\
\bottomrule
\end{tabular}%
}
\end{table}

\end{document}